\documentclass[10pt,twocolumn,letterpaper]{article}

\usepackage{cvpr}              

\usepackage[dvipsnames]{xcolor}

\usepackage[accsupp]{axessibility}

\usepackage{algorithm}
\usepackage{algpseudocode}
\usepackage{multirow}
\definecolor{cvprblue}{rgb}{0.21,0.49,0.74}
\usepackage[pagebackref,breaklinks,colorlinks,citecolor=cvprblue]{hyperref}

\newcommand{\argmin}{\operatorname*{argmin}}

\def\paperID{} 
\def\confName{CVPR}
\def\confYear{2024}

\title{Resurrecting Old Classes with New Data for Exemplar-Free Continual Learning}

\author{Dipam Goswami\textsuperscript{1,2} \enspace Albin Soutif--Cormerais\textsuperscript{1,2} \enspace  Yuyang Liu\textsuperscript{3,4,$\dagger$} \enspace Sandesh Kamath\textsuperscript{1,2} \\ 
\enspace Bartłomiej Twardowski\textsuperscript{1,2,5} \enspace Joost van de Weijer\textsuperscript{1,2}  \\ 
\\
\textsuperscript{1}Department of Computer Science, Universitat Autònoma de Barcelona \\
\textsuperscript{2}Computer Vision Center, Barcelona \space
\textsuperscript{3}University of Chinese Academy of Sciences \\
\textsuperscript{4}Shenyang Institute of Automation, Chinese Academy of Sciences \space
\textsuperscript{5}IDEAS-NCBR \\  
{\tt\small \{dgoswami, albin, skamath, btwardowski, joost\}@cvc.uab.es, liuyuyang@sia.cn}
}

\begin{document}
\maketitle
\begin{abstract}

Continual learning methods are known to suffer from catastrophic forgetting, a phenomenon that is particularly hard to counter for methods that do not store exemplars of previous tasks. Therefore, to reduce potential drift in the feature extractor, existing exemplar-free methods are typically evaluated in settings where the first task is significantly larger than subsequent tasks. Their performance drops drastically in more challenging settings starting with a smaller first task. To address this problem of feature drift estimation for exemplar-free methods, we propose to adversarially perturb the current samples such that their embeddings are close to the old class prototypes in the old model embedding space. We then estimate the drift in the embedding space from the old to the new model using the perturbed images and compensate the prototypes accordingly. We exploit the fact that adversarial samples are transferable from the old to the new feature space in a continual learning setting. The generation of these images is simple and computationally cheap. We demonstrate in our experiments that the proposed approach better tracks the movement of prototypes in embedding space and outperforms existing methods on several standard continual learning benchmarks as well as on fine-grained datasets. Code is available at \url{https://github.com/dipamgoswami/ADC}.
\end{abstract}    

\let
\thefootnote\relax
\footnotetext{\textsuperscript{$\dagger$} Yuyang Liu is the corresponding author.}

\section{Introduction}
\label{sec:intro}

Deep learning has gained widespread use in various computer vision tasks, demonstrating exceptional performance when trained on a dataset in a single session. However, a significant challenge arises when new data is introduced incrementally in multiple phases or tasks. Then neural networks need to adapt without forgetting previously learned information, a phenomenon known as \textit{catastrophic forgetting}~\cite{mccloskey1989catastrophic,kemker2018measuring}.
Recent studies in continual learning (CL)~\cite{de2021continual, masana2020class, zhou2023deep, wang2023comprehensive} focus on two prevalent scenarios~\cite{van2019three}: Task-Incremental Learning (TIL), where task information is available during testing, and Class-Incremental Learning (CIL), where it is not. Our work aims to address the more challenging CIL problem. 

\begin{figure}[t]
\centering
\includegraphics[width=1\linewidth]{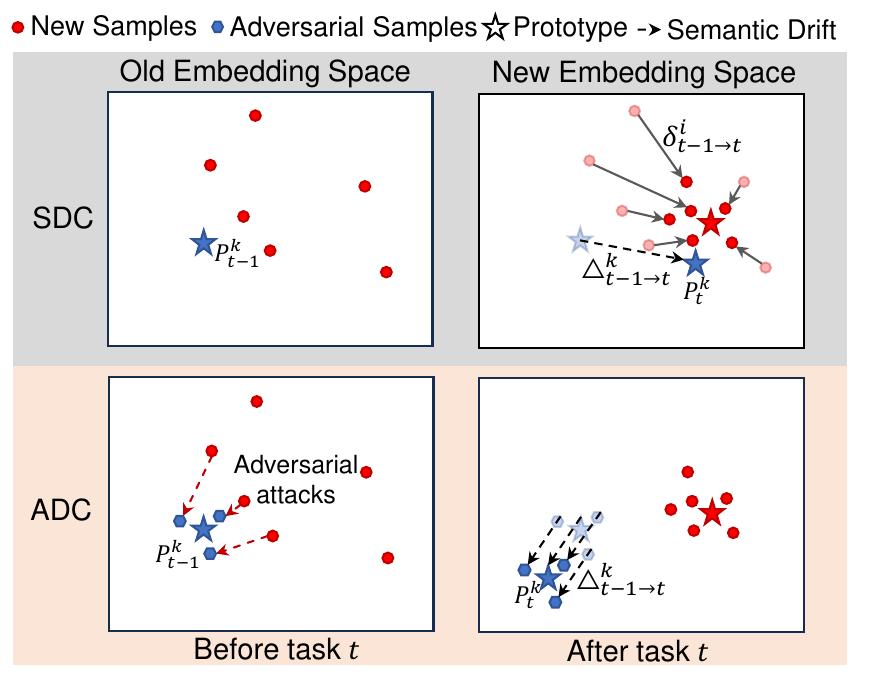}
\caption{Illustration of Adversarial Drift Compensation (ADC) and SDC~\cite{Yu_2020_CVPR}. In SDC, the drift $\Delta^k_{t-1 \xrightarrow{} t}$ is estimated as the average of drift of all new task samples after training on a new task. Instead, we propose to move the new task features close to the old prototype $P^k_{t-1}$ of class $k$ by perturbing the new images using targeted adversarial attacks. The drift of the adversarial samples from old to new feature space is used to resurrect all old prototypes.} 
\label{fig:intro}
\vspace{-12pt}
\end{figure}

Exemplar-based CIL methods~\cite{belouadah2019il2m, hou2019learning, dhar2019learning, douillard2020podnet, chaudhry2018riemannian, rolnick2019experience,liu2023augmented,wang2022foster} store small subsets of data from each task. These exemplars are later replayed with current data during training in new tasks. 
Although effective, these methods necessitate storing input data from previous tasks, leading to multiple challenges in practical settings such as legal concerns with new regulations (e.g. European GDPR where users can request to delete personal data), and privacy issues when dealing with sensitive data like in medical imaging.
Recently, the \emph{exemplar-free CIL} (EFCIL) setting is extensively studied ~\cite{petit2023fetril,zhu2022self,zhu2021prototype,goswami2023fecam,Yu_2020_CVPR,Malepathirana2023napa,Zhu2023self}. However, unlike exemplar-based methods, the EFCIL methods are only effective when starting with high-quality feature representations and are thus dependent on having a large initial task which is typically half of the whole dataset. However, a more practical CIL approach should be able to perform well on training from a smaller initial task and at the same time should not store exemplars. We define this as a \emph{small-start setting} and analyze how existing EFCIL methods perform in this setting.

A critical aspect in CIL is the \textit{semantic drift} of feature representations~\cite{Yu_2020_CVPR} after training on new tasks. This results in the movement of class distributions in feature space. Thus, it is crucial to track the old class representations after learning new tasks.
While the class-mean in the new feature space can be effectively estimated using Nearest-Mean of Exemplars (NME)~\cite{douillard2020podnet,rebuffi2017icarl}, it is challenging to estimate it without exemplars.
Usually, this drift is minimized with heavy functional regularization, which consequently restricts the plasticity of the network. Another way 
is to estimate it from the drift of current data, as done in SDC~\cite{Yu_2020_CVPR} or by augmenting old prototypes using new class features~\cite{Shi2023prototype,Malepathirana2023napa}. In this paper, we propose a novel drift estimation method using adversarial examples to resurrect old class prototypes in the new feature space as shown in~\cref{fig:intro}.

Adversarial examples~\cite{szegedy2014intriguing, athalye2018synthesizing, ilyas2019adversarial, madry2018towards, moosavi2016deepfool} are maliciously crafted inputs that are designed to fool a neural network into predicting a different output than the one initially predicted for the original input. Exploiting the concept of targeted adversarial attacks~\cite{kurakin2016adversarial,madry2018towards}, we propose to perturb the new data such that the adversarial images result in embeddings close to the old prototypes. Now, the drift from old to new feature space is estimated using these adversarial samples, which serve as pseudo-exemplars for the old classes.
We hypothesize that the pseudo-exemplars behave like the original exemplars in the feature space, and thus we exploit them to measure the drift.
This generation of adversarial samples is computationally cheaper and much faster (only a few iterations) compared to data-inversion methods~\cite{yin2020dreaming} which inverts embeddings to realistic images.

Following recent studies~\cite{janson2022simple,goswami2023fecam,ma2023progressive}, we explore using class prototypes with an NCM~\cite{rebuffi2017icarl} classifier and show that a simple baseline of logits distillation~\cite{li2017learning} with an NCM classifier often outperforms existing EFCIL methods in the small-start setting. Applying our proposed drift compensation method with this baseline, we obtain state-of-the-art performance with significant gains over existing methods on standard CL benchmarks using CIFAR-100~\cite{krizhevsky2009learning}, TinyImageNet~\cite{le2015tiny} and ImageNet-Subset~\cite{deng2009imagenet} as well as fine-grained datasets like CUB-200~\cite{wah2011caltech} and Stanford Cars~\cite{krause20133d}.
Our contributions can be summarized as:
\begin{itemize}
    \item We study the challenging EFCIL settings and highlight the importance of continually learning from small-start settings instead of assuming the availability of half of the dataset in the first task.
    \item We present a novel and intuitive method - Adversarial Drift Compensation (ADC) to estimate semantic drift and resurrect old class prototypes in the new feature space. We also investigate how adversarially generated samples transfer in CIL settings from old to the new model.
    \item We perform experiments on several CIL benchmarks and outperform state-of-the-art methods by a large margin on several benchmark datasets. Especially notable are our results on fine-grained datasets, where we report performance gains of around 9\% for last task accuracy. 
\end{itemize}

\section{Related Work}
\label{sec:related}

\textbf{Class-Incremental Learning.} CIL~\cite{masana2020class,zhou2023deep,de2021continual} methods aim to learn new data which arrives incrementally and suffers from the catastrophic forgetting problem~\cite{robins1995catastrophic,mccloskey1989catastrophic}. During evaluation in CIL without the task id, it is difficult to distinguish classes that belong to different tasks~\cite{soutif2021importance}. 
While in general this setting is tackled using \textit{rehearsal approaches}~\cite{belouadah2019il2m, hou2019learning, dhar2019learning, douillard2020podnet, rolnick2019experience} by storing raw inputs, some attempts have been made without storing raw inputs. LwF \cite{li2017learning} prevents important changes in the network by preventing the output of the current model to drift too much from the output of the previous model.
PASS~\cite{zhu2021prototype} learns the backbone using self-supervised learning and later uses functional regularization and feature rehearsal, SSRE~\cite{zhu2022self} proposed an architecture organization strategy that aims to transfer invariant knowledge across tasks. In FeTRIL~\cite{petit2023fetril}, the authors freeze the feature extractor and estimate the position of the old class features by using the current task data variance. Recently, FeCAM~\cite{goswami2023fecam} leveraged the mean and covariance of the previous task features and proposed a  mahalanobis distance-based classifier.

\noindent\textbf{Drift estimation.} When updating the feature extractor on new classes, the representation learned for the old class prototypes changes and thus the need to rectify those drifts~\cite{Yu_2020_CVPR}. SDC~\cite{Yu_2020_CVPR} showed that the new data can be used to estimate the drift of the old prototype representations. Recent methods~\cite{Toldo2022bring,Malepathirana2023napa,Shi2023prototype} also explored how to update the prototypes learned in old tasks to counter the drift. Toldo et al. proposed to learn the relations between old and new class features to estimate the drift. NAPA-VQ~\cite{Malepathirana2023napa} proposed to augment the prototypes using the topological information of classes in the feature space. Prototype Reminiscence~\cite{Shi2023prototype} proposed to dynamically reshape old class feature distributions by interpolating the old prototypes with the new sample features. In this work, we generate adversarial samples which behaves as pseudo-exemplars and is then used to measure the drift.

\noindent\textbf{CL using Adversarial Attacks.} 
Adversarial Attacks has been studied in-depth in recent years~\cite{szegedy2014intriguing, athalye2018synthesizing, ilyas2019adversarial}, and has been later harnessed to create realistic looking images from a trained vision model~\cite{mordvintsev2015inceptionism}, including inputs that can be later used for training~\cite{yin2020dreaming}. 
Some recent methods~\cite{ebrahimi2020adversarial,shim2021online,jin2021gradient,kumari2022retrospective} in exemplar-based CIL borrowed the idea of adversarial attacks. ASER~\cite{shim2021online} used the kNN-specific Shapley value to obtain more representative buffer samples. GMED~\cite{jin2021gradient} edits the exemplars by monitoring the change in loss when training on incoming data. RAR~\cite{kumari2022retrospective} used the pairwise relations between the exemplars and the new samples and perturb the exemplars to obtain samples close to the decision boundaries. While all these approaches use adversarial attacks on the memory samples, we use it to perturb the new data to simulate the old data.

\section{Method}
\label{sec:method}
We consider the EFCIL setup where new classes emerge over time and we are not allowed to store samples from old classes. These classes come in different tasks, one task at a time, and the tasks contain a mutually exclusive set of classes. When training on task $t$, we have access to current dataset $D_t = \{X_t, Y_t \}$ with images $X_t$ and labels $Y_t$. The main goal of EFCIL is to learn a model $h$ that correctly classifies the data into classes encountered so far. We use $h_t(x) = \sigma(W_t f_t(x))$, where $f_t$ is the feature extractor parameterized by $\theta_t$ learned in task $t$ and $W_t$ is weight matrix of the linear classifier with softmax function $\sigma$.

\subsection{Motivation}
In general, for a new feature extractor $f_t$ trained on new data and an old feature extractor $f_{t-1}$ from previous task, we have access to old class prototypes up to task $t-1$ denoted by $P_{t-1}^{Y_{1:t-1}}$. We compute the prototypes for all new classes after training in the current task. For a class $k$ in task $t$, we compute 
$P_{t}^k = \frac{1}{|X_t^k|}\sum_{x\in X_t^k} f_{t}(x)$, where $X_{t}^k$ is the set of samples from class $k$, $f_{t}(x)$ is the feature embedding for an image $x$ from class $k$. 
However, the old class prototypes were computed on the old feature space $f_{t'}$ ($t' < t$) in old tasks and have drifted to a different position in the new feature space $f_t$ after training on new data.

Previously, SDC~\cite{Yu_2020_CVPR} proposed to compensate the drift of these prototypes $P^{Y_{1:t-1}}_{t-1}$ by computing the drift from old model embeddings $f_{t-1}(x)$ to new model embeddings $f_t(x)$ corresponding to all images $x$ in the current task. This drift of the current data is then used to approximate the drift of previous task prototypes by considering a weighted Gaussian window around the prototypes (giving more weights to drift vectors close to the prototype). However, the quality of this drift approximation for previous prototypes is expected to be low when few current task data points are close to a given prototype in the embedding space.
We show in our experiments that SDC indeed struggles to estimate the drift in the small-start settings where the feature representations change considerably.

\begin{figure}[t]
    \centering
    \includegraphics[trim=0 0 0 0, clip, width=0.95\linewidth]{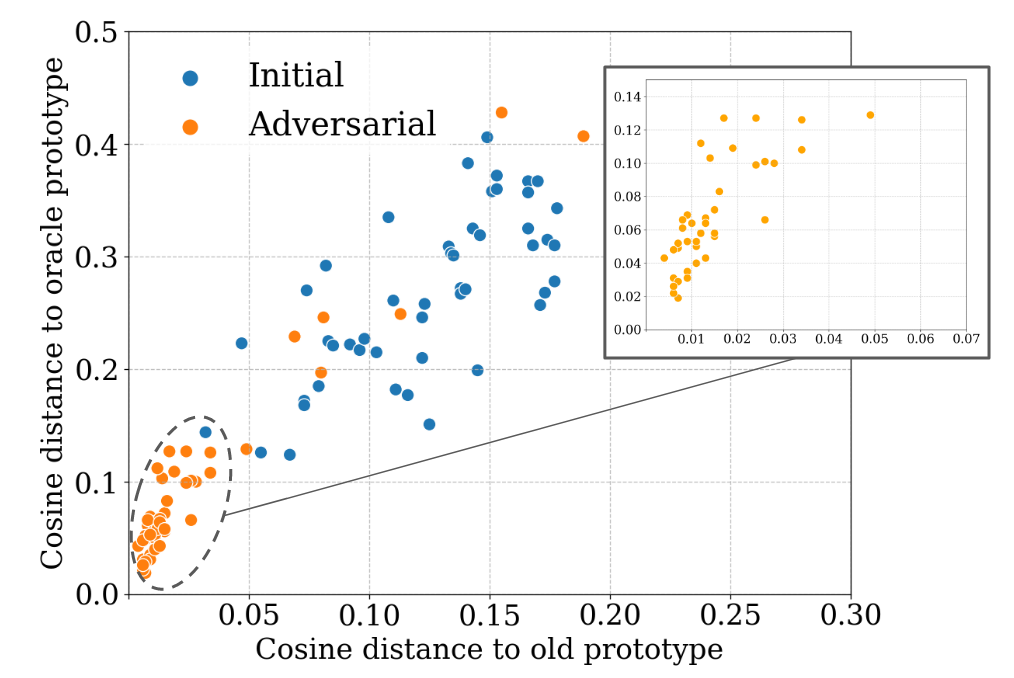}
    \caption{Illustration to show that the cosine distance between embeddings and old prototype in the old feature space is correlated with the cosine distance between embeddings and oracle prototype in the new feature space. This holds true for embeddings of both initial and adversarial samples.
    For demonstration, we select few current-task samples that are closest to the old prototype and choose the same target old class for all samples. The blue and orange points represents the non-modified current class samples and the modified samples using our proposed approach respectively. In this analysis, we compute the oracle prototype using all old task data in the new feature space.}
    \label{fig:motivation}
    \vspace{-12pt}
\end{figure}

In a similar fashion, it is possible to estimate this drift without using such weights, by simply choosing the closest samples to each old class prototype and compute the average feature of these samples when fed to the new backbone. 
We select samples from the current task that are close to the old prototype of a given class and verify that such samples also lie close to the oracle prototype (in the new feature space).
We analyze in Fig.~\ref{fig:motivation} that there exist a correlation between the distance to the old prototype in the old feature space and the distance to the oracle prototype in the new feature space (see blue dots). This motivates us to leverage current task samples so that their distance to the old prototype in the old feature space is even smaller, which could in turn improve the drift estimation. We hypothesize that this can be done by computing adversarial samples from the current task samples, aiming for their representation to match one of the old class prototypes in the old feature space.

\begin{figure*}[t]
\centering
\includegraphics[trim=0 0 0 0, clip,width=\linewidth]{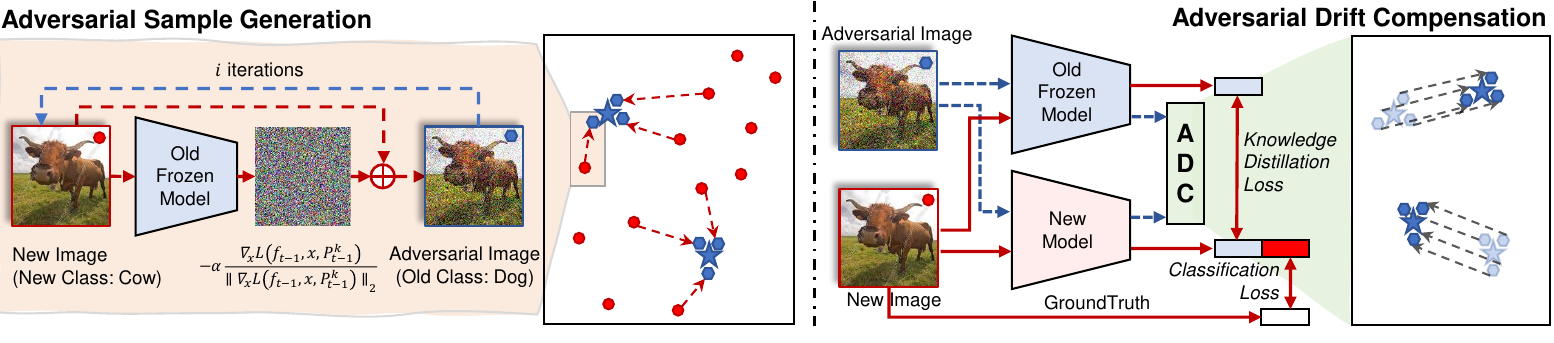}
\caption{(a) Adversarial Sample Generation: On the old model feature space, the new samples closest to the old prototype are selected and iteratively perturbed in the direction of the target old prototype to generate adversarial samples which are now misclassified as the target old class resulting in embeddings closer to the old prototype. We perform this for every old class (we show 2 classes here for demonstration). (b) Model Training with Drift Compensation: The new model is trained using the classification loss for learning new classes and knowledge distillation loss to prevent forgetting of old classes. After the new model is trained, the adversarial samples generated using the old model are passed through both the models and the drift from old to new feature space is estimated. This is then used to update the old prototypes. } 
\label{fig:ADC}
\vspace{-8pt}
\end{figure*}

\subsection{Adversarial Drift Estimation}
To estimate the drift of old class prototypes after updating the model on new classes, it is desirable to have the exemplars.
These exemplars can be passed through the new model to compute the \textit{oracle} prototype position in the new feature space.
However, in the exemplar-free setting, we can only access the new data. In order to use the new data to represent the old data, we exploit the concept of targeted adversarial attacks~\cite{kurakin2016adversarial,madry2018towards} to target one old class at a time and perturb the new data in a way that it serves as a substitute of old data to the model. We perform adversarial attacks on new data to move its embeddings very close to old prototypes in the old feature space. 
Using the adversarial samples, we can estimate the drift from old to new feature space and compensate it as illustrated in~\cref{fig:ADC}.

To estimate the drift of prototype $P^k_{t-1}$ 
for a target old class $k$, we obtain $\mathcal{X}^k$ by sampling a set of $m$ data points from the current task data $X_t$ which are closest to $P^k_{t-1}$ based on L2 distance between the embeddings of samples in $X_t$ and the prototype $P^k_{t-1}$.
We aim to perturb the samples $x \in \mathcal{X}^k$ and obtain $\mathcal{X}^k_{adv}$ such that the adversarial samples $x_{adv} \in \mathcal{X}^k_{adv}$ are closer to $P^k_{t-1}$ and are now classified to class $k$ using the NCM classifier in the old feature space:
\begin{align}
    k = \argmin\limits_{y \, \in \, Y_{1:t-1}}  \| f_{t-1}(x_{adv}) - P^y_{t-1} \|_2.
    \label{ncm}
\end{align}

We propose the following optimization objective by computing the mean squared error between the features $f_{t-1}(x)$ and the prototype $P^k_{t-1}$ as:
\begin{align}
    L(f_{t-1},\mathcal{X}^k,P^k_{t-1}) = \frac{1}{|\mathcal{X}^k|} \sum_{x \, \in \, \mathcal{X}^k} \|f_{t-1}(x) - P^k_{t-1}\|_2^2.
\label{adv_loss}
\end{align}
In order to move the feature embedding in the direction of the target prototype $P^k_{t-1}$,
we obtain the gradient of the loss with respect to the data $x \in \mathcal{X}^k$, normalize it to get the unit attack vector and scale it by $\alpha$ as follows:
\begin{align}
    x_{adv} \xleftarrow{} x - \alpha \frac{\nabla_{x} L(f_{t-1},x,P^k_{t-1})}{\|\nabla_{x} L(f_{t-1},x,P^k_{t-1})\|_2} \;\forall x \in \mathcal{X}^k
\end{align}
where $\nabla_{x} L(f_{t-1},x,P^k_{t-1})$ is the gradient of the objective function with respect to the data $x$ and $\alpha$ refers to the step size. We perform the attack for $i$ iterations.

Here, the goal is different from conventional adversarial attacks like FGSM and its variants~\cite{goodfellow2014explaining,dong2018boosting,kurakin2016adversarial,madry2018towards} which aim to minimize the perturbation in order to keep the perturbed image visually similar to the real image by having a fixed $\epsilon$-budget generally based on $\ell_{2}$ or $\ell_{\infty}$-norm of perturbation. In our case, we do not need to apply such restrictions on the distance between initial and final image, instead, we only clip the perturbed image in the existing range of pixel values. We show in supplementary materials that indeed the generated adversarial images have much higher perturbation.
We do observe that our formulation is closer to the $\ell_{2}$-norm based attack as we use $\ell_{2}$ normalization of the gradient vector to obtain a unit perturbation vector which is scaled using the step size.

\noindent\textbf{Continual Adversarial Transferability.} 
An interesting aspect of our method is that the adversarial samples $x_{adv}$ are crafted on the old feature extractor $f_{t-1}$ and then passed to the new feature extractor $f_t$ expecting that the adversarial samples will still be misclassified as the target old class $k$. We define this as \textit{continual adversarial transferability} where the adversarial samples generated on the old feature space still behave in the same way on the new feature space. This is feasible since the old model and the new model are not entirely different because of the knowledge distillation used in order to reduce catastrophic forgetting. This is related with the concept of adversarial transferability~\cite{papernot2016transferability,moosavi2017universal,madry2018towards,inkawhich2019feature}, where an attack obtained on one neural network also behaves as an attack on other independently trained neural network based architectures. 

We analyze the oracle setting using the old class data to validate the continual adversarial transferability. We show in Fig.~\ref{fig:motivation} that distance of the adversarial samples from their target prototypes in the old feature space is still correlated with their distance to the oracle prototypes of the target class in the new feature space. This suggests that the adversarial samples crafted using the old feature space are still effective in the new feature space and therefore allows us to reliably compute the drift from these adversarial samples.

\subsection{Drift Compensation}
The adversarial samples when passed through the new feature extractor $f_t$ are expected to lie close to the drifted prototype and hence are used to compute the drift. After generating the adversarial samples for each target class $k$, we measure the prototype drift as:
\begin{align}
    \Delta^k_{t-1 \xrightarrow{} t} = \frac{1}{|\mathcal{X}^k_{adv}|} \sum_{x_{adv} \in \mathcal{X}^k_{adv}} (f_t(x_{adv}) - f_{t-1}(x_{adv}))
    \label{compensation}
\end{align}
where $x_{adv} \in \mathcal{X}^k_{adv}$ is the set of only those adversarial samples which are classified as the target class $k$ using the NCM classifier.
We resurrect the old prototypes by compensating the drift as follows:
\begin{align}
    P^k_t = P^k_{t-1} + \Delta^k_{t-1 \xrightarrow{} t}
\end{align}
After compensating all old prototypes, we use the NCM classifier in the new feature space for classifying the test samples. 
Unlike SDC~\cite{Yu_2020_CVPR}, we do not perform weighted averaging based on the distances to the prototype since embeddings from adversarial images are very close to the prototypes and we found no gain by applying this additional weighting scheme.

\subsection{Training Strategy}
In addition to learning new classes, we perform knowledge distillation~\cite{li2017learning} on the logits to transfer knowledge from the frozen teacher model at previous task $t-1$ to the student model at current task $t$ as follows: 

\begin{align}
    \mathcal{L} = \mathcal{L}_{ce}(h_t(x), y) + \lambda \mathcal{L}_{ce}(h_{t-1}(x), h_t(x))
\end{align}
where $\lambda$ refers to the regularization strength, the first term $\mathcal{L}_{ce}$ refers to the cross-entropy loss for learning new classes and the second term performs the regularization by forcing the probabilities of old classes on the old model $h_{t-1}$ and new model $h_t$ to be similar and thus prevents forgetting.

\begin{algorithm}
\caption{Adversarial Drift Compensation}
\begin{algorithmic}[1]
\Statex \textbf{input:} Images $X_t$, feature extractors $f_{t}$ and $f_{t-1}$, old prototypes $P^{Y_{1:t-1}}_{t-1}$, step size $\alpha$, number of generated samples $m$, number of iterations for perturbation $i$.
\Statex \textbf{output:} Resurrected prototypes $P^{Y_{1:t-1}}_t$.
\For{$k \, \in \, {Y_{1:t-1}}$}
\State Sample a set $\mathcal{X}^k$ of $m$ new samples from $X_t$ which are closest to $P^k_{t-1}$ based on L2 distance.
\For{$i$ iterations}
\State $L(f_{t-1},\mathcal{X}^k,P^k_{t-1}) =$
\Statex \hspace{80pt} $ \frac{1}{|\mathcal{X}^k|} \sum_{x \in \mathcal{X}^k} \|f_{t-1}(x) - P^k_{t-1}\|_2^2$.
\State $x_{adv} = x - \alpha \frac{\nabla_{x} L(f_{t-1},x,P^k_{t-1})}{\|\nabla_{x} L(f_{t-1},x,P^k_{t-1})\|_2} \forall x \in \mathcal{X}^k$ 
\State $x \gets x_{adv}$
\EndFor

\State add $x$ in $\mathcal{X}^k_{adv}$ if $\argmin\limits_{y \, \in \, Y_{1:t-1}} \| f_{t-1}(x) - P^y_{t-1} \|_2 = k$.
\Statex \Comment{Store only those samples in $\mathcal{X}^k_{adv}$ which are successfully misclassified as k}
\State $\Delta^k_{t-1 \xrightarrow{} t} = \frac{1}{|\mathcal{X}^k_{adv}|}\sum\limits_{x_{adv} \in \mathcal{X}^k_{adv}}f_t(x_{adv})-f_{t-1}(x_{adv})$
\State $P^k_t = P^k_{t-1} + \Delta^k_{t-1 \xrightarrow{} t}$ \Comment{Prototype resurrection}
\EndFor
\end{algorithmic}
\label{alg:adc}
\end{algorithm}

\section{Experiments}

\noindent \textbf{Datasets.} We perform experiments on several CIL benchmarks. CIFAR-100 \cite{krizhevsky2009learning} contains 50k training images of size 32x32 and 10k test images, divided in 100 classes. TinyImageNet\cite{le2015tiny} contains 100k training images and 10k test images from 200 classes and image size of 64x64, taken as a subset of ImageNet \cite{deng2009imagenet}. ImageNet-Subset is a subset of the ImageNet~(ILSVRC 2012) dataset ~\cite{russakovsky2015imagenet} containing 100 classes with a total of 130k training images and 5k test images and image size of 224x224. 
We equally split all these datasets in 5 and 10 tasks.
This is different from the big-start settings with half of the dataset in first task, commonly used in EFCIL benchmarks~\cite{goswami2023fecam,petit2023fetril,zhu2021prototype}.
We also use two fine-grained datasets for our experiments. 
CUB-200~\cite{wah2011caltech} contains 200 classes of birds with 224x224 image size, 5994 images for training and 5794 images for testing. We use the 5-split and 10-split settings for CUB-200. 
Stanford Cars~\cite{krause20133d} consists of 196 car models with 224x224 images, 8144 for training and 8041 for testing and we split it into 7 and 14 tasks.

\begin{table*}[t]
\begin{center}
\resizebox{0.95\textwidth}{!}{
\begin{tabular}{lcccccccccccc}
\toprule
\multirow{3}{*}{Method}    & \multicolumn{4}{c}{CIFAR-100}                        & \multicolumn{4}{c}{TinyImageNet}                      & \multicolumn{4}{c}{ImageNet-Subset}                    \\
 \cmidrule(lr){2-5} \cmidrule(lr){6-9} \cmidrule(l){10-13}
           & \multicolumn{2}{c}{T = 5} & \multicolumn{2}{c}{T=10} & \multicolumn{2}{c}{T = 5} & \multicolumn{2}{c}{T =10} & \multicolumn{2}{c}{T = 5} & \multicolumn{2}{c}{T = 10} \\
           & $A_{last}$     & $A_{inc}$    & $A_{last}$    & $A_{inc}$    & $A_{last}$     & $A_{inc}$    & $A_{last}$     & $A_{inc}$    & $A_{last}$     & $A_{inc}$    & $A_{last}$     & $A_{inc}$     \\
           \midrule
LwF~\cite{li2017learning}        &      45.35        &     61.94       &      26.14       &     46.14       &       38.81       &     49.70       &      \underline{27.42}        &    38.77        &      50.88        &      69.11      &      37.90        &     61.60       \\
NCM       &      53.53        &     \underline{66.35}       &      41.31       &     57.85       &           38.69     &       50.45       &     26.56      &     \underline{41.04}      &      57.74        &     71.99       &        \underline{45.86}      &      65.04       \\
SDC~\cite{Yu_2020_CVPR}        &      \underline{54.94}        &      64.82      &      \underline{41.36}       &    \underline{58.02}        &       \underline{40.05}       &      \underline{50.82}      &       27.15       &      40.46      &       \underline{59.82}       &      \underline{74.10}      &      43.72        &      \underline{65.83}       \\
PASS~\cite{zhu2021prototype}       &       49.75       &     63.39       &     37.78        &     52.18       &       36.44       &    48.64        &       26.58       &      38.65      &       50.96       &      66.15      &       38.90       &     54.74        \\
SSRE~\cite{zhu2022self} &       42.39       &    56.57     &       29.44      &    44.38      &        30.13      &      43.20      &        22.48      &    34.93        &      40.30        &      57.57      &       28.12       &      45.87       \\
FeTrIL~\cite{petit2023fetril}     &      45.11        &      60.42      &      36.69       &     52.11       &       29.91       &      43.99      &      23.88        &     36.35       &        49.18      &    63.83        &       40.26       &       55.12      \\
FeCAM~\cite{goswami2023fecam}      &       47.28       &      61.37      &       33.82      &      48.58      &       25.62       &     39.85       &       23.21       &    35.32        &       54.18       &      67.21      &       42.68       &      57.45       \\
\midrule
ADC (Ours) &     \textbf{59.14}         &     \textbf{69.62}       &     \textbf{46.48}        &     \textbf{61.35}       &     \textbf{41.0}         &      \textbf{50.94}      &     \textbf{32.32}         &    \textbf{43.04}        &       \textbf{62.40}       &    \textbf{74.84}        &        \textbf{47.58}      &     \textbf{67.07}    \\
\bottomrule

\end{tabular}
}
\caption{Evaluation of EFCIL methods on small-start settings. Best results in \textbf{bold} and second best results are \underline{underlined}.}
\label{table1}
\end{center}
\vspace{-12pt}
\end{table*}

\noindent \textbf{Training Details.} We use the PyCIL framework~\cite{zhou2023pycil} as a basis for all our experiments. The training is performed using the ResNet18 model~\cite{he2016deep} and the SGD optimizer. 
For CIFAR-100, in the first task, we use a starting learning rate of 0.1, momentum of 0.9, batch size of 128 and weight decay of 5e-4 for 200 epochs, the learning rate is reduced by a factor of 10 after 60, 120, and 160 epochs. In the subsequent tasks, we use an initial learning rate of 0.05 reduced by a factor of 10 after 45 and 90 epochs and train for 100 epochs. Following~\cite{masana2020class}, we set the regularization strength to 10 and the temperature to 2. 
The network is trained from scratch on CIFAR-100, TinyImageNet and ImageNet-Subset.
For the experiments on fine-grained datasets, we use the ImageNet pretrained weights following standard practice~\cite{Yu_2020_CVPR,rymarczyk2023icicle}. For ADC, we use a $\alpha$ value of 25, iterations $i = 3$ and number of closest samples $m=100$ for all the datasets. Similar to most existing methods, we store all the class prototypes. Complete details about the training setting for all the datasets are given in the supplementary materials.

\noindent \textbf{Compared Methods.} 
Since none of the EFCIL methods are designed to start from a small first task, we implement those methods in our small-start settings. This includes LwF~\cite{li2017learning}, PASS~\cite{zhu2021prototype}, SSRE~\cite{zhu2022self}, FeTRIL~\cite{petit2023fetril} and FeCAM~\cite{goswami2023fecam}. 
Naturally, we also include a comparison to the existing drift-estimation method SDC~\cite{Yu_2020_CVPR} and the baseline model with NCM classifier. For SDC and NCM results reported in~\cref{table1} and~\cref{table2}, we train the models using LwF and perform NCM classification in the feature space. For FeTrIL and FeCAM, the feature extractor is frozen after the first task, while for the other methods, it is continually learned. Note that here we adapt SDC with distillation on the logits, which is different from~\cite{Yu_2020_CVPR} where they performed distillation on the features.

\begin{table*}[t]
\begin{center}

\resizebox{0.7\textwidth}{!}{
\begin{tabular}{lcccccccc}
\toprule
\multirow{3}{*}{Method}    & \multicolumn{4}{c}{CUB-200}                        & \multicolumn{4}{c}{Stanford Cars}                   \\
 \cmidrule(lr){2-5} \cmidrule(lr){6-9}
           & \multicolumn{2}{c}{T = 5} & \multicolumn{2}{c}{T=10} & \multicolumn{2}{c}{T = 7} & \multicolumn{2}{c}{T =14} \\
           & $A_{last}$     & $A_{inc}$    & $A_{last}$    & $A_{inc}$    & $A_{last}$     & $A_{inc}$    & $A_{last}$     & $A_{inc}$      \\
           \midrule
LwF~\cite{li2017learning}         &      \underline{58.68}        &     \underline{71.31}      &      41.96       &    60.15        &        \underline{45.18}      &     61.14       &      30.33        &      49.93   \\
NCM        &      52.74        &     67.13       &      38.47       &     57.83       &      42.22        &      59.06      &      31.60        &     51.34     \\
SDC~\cite{Yu_2020_CVPR}        &      55.20        &      68.64      &      41.63       &     60.43       &      45.03        &     \underline{61.75}       &       32.15       &      \underline{53.18}        \\
PASS~\cite{zhu2021prototype}       &      34.04        &      49.00      &       26.37      &      41.08      &       20.71       &      37.13      &        12.30      &     25.46       \\
FeTrIL~\cite{petit2023fetril}     &      54.66        &      67.45      &      49.09       &     62.42       &       36.92       &     54.09       &      34.29        &     50.41       \\
FeCAM~\cite{goswami2023fecam}      &      53.47        &      66.39      &      \underline{51.78}       &     \underline{64.97}       &       40.64       &     56.24       &       \underline{37.50}       &      52.78         \\
\midrule
ADC (Ours) &      \textbf{64.46}        &       \textbf{73.49}     &      \textbf{57.97}       &      \textbf{68.91}      &      \textbf{54.86}        &     \textbf{67.07}       &     \textbf{45.07}         &    \textbf{61.39}       \\
\bottomrule

\end{tabular}
}
\caption{Evaluation of EFCIL methods on fine-grained datasets. Best results in \textbf{bold} and second best results are \underline{underlined}.}
\label{table2}
\end{center}
\vspace{-12pt}
\end{table*}

\noindent \textbf{Evaluation.} We report the average accuracy after the last task denoted by $A_{last}$ and the average incremental accuracy which is the average of the accuracy after all tasks (including the first one) denoted by $A_{inc}$. $A_{inc}$ better reflects the performance of the methods across all the tasks.

\subsection{Quantitative Evaluation}
We observe that methods proposed for the big-start settings of EFCIL are not effective in small-start settings and perform poorly. A simple baseline trained with LwF and using NCM classifier is performing better than most of the existing approaches - SSRE, PASS, FeTrIL and FeCAM in several settings. While SDC improves over NCM, the proposed method ADC outperforms all existing methods in both last task accuracy and average incremental accuracy across all settings in~\cref{table1} and~\cref{table2}. ADC outperforms the second-best method SDC by 4.2\% on 5-task and by 5.12\% on 10-task settings of CIFAR-100 on last-task accuracy. For TinyImageNet, ADC improves over the second-best method by 0.95\% on 5-task and by 5.17\% on 10-task settings. On ImageNet-Subset, ADC is better by 2.58\% on 5-task and by 1.72\% on 10-task settings after the last task.

\begin{figure*}[t]
\centering
\includegraphics[width=1.0\linewidth]{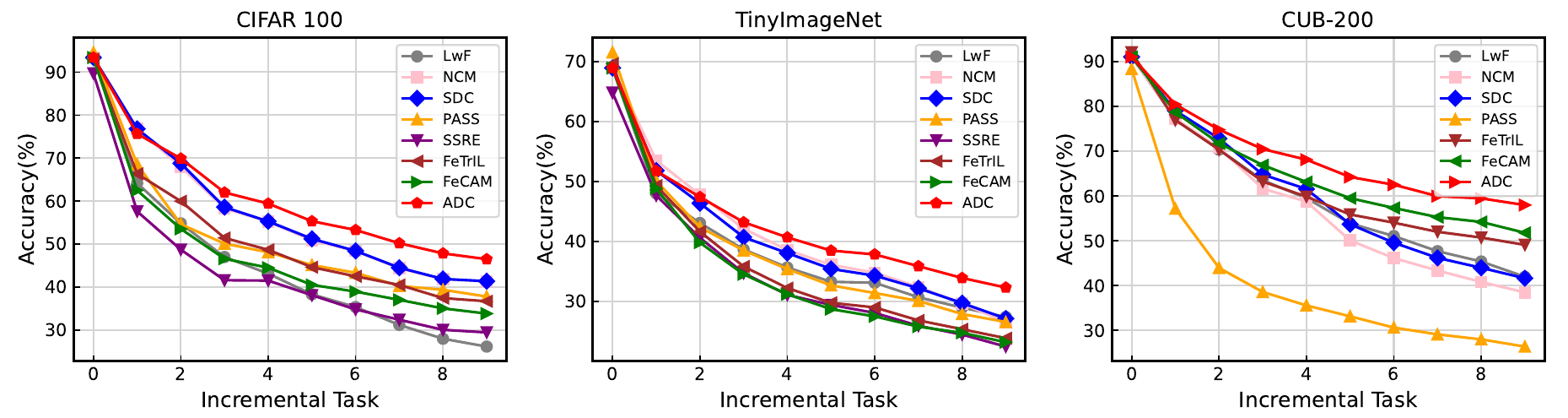}
\caption{Accuracy after each incremental task for CIFAR-100, TinyImageNet and CUB-200 datasets on 10 task settings. ADC improves over the compared methods starting from the initial to the last task.}
\label{fig:inc}
\vspace{-12pt}
\end{figure*}

We also evaluate the EFCIL methods on the challenging fine-grained datasets of CUB-200 and Stanford Cars. We observe in~\cref{table2} that LwF is a strong baseline here, particularly in the 5-task and 7-task settings and methods like NCM and SDC are not much better than LwF. While PASS performs poorly on both datasets, FeTrIL and FeCAM performs better with FeCAM outperforming the other methods on the 10-task setting of CUB-200 and 14-task setting of Stanford Cars. ADC outperforms the runner-up methods by 5.78\% on 5-task setting and by 6.19\% on 10-task settings of CUB-200. On Stanford Cars dataset, ADC is better by 9.68\% on 7-task setting and 7.57 \% on 14-task setting. We analyze how the accuracy after each task varies for all the methods in~\cref{fig:inc} and observe that ADC consistently outperforms the other methods across all tasks.

\begin{figure}[t]
\centering
\includegraphics[width=0.9\linewidth]{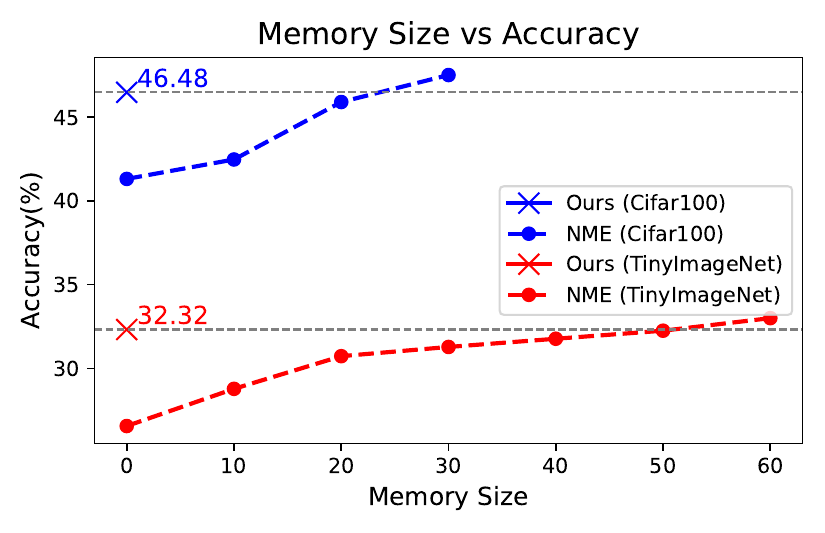}
\caption{Memory Size vs accuracy comparison of NME and ADC on CIFAR-100 and TinyImageNet (T=10) settings. }
\label{fig:mem}
\vspace{-15pt}
\end{figure}
\noindent \textbf{Comparison to NME.} We compare the last-task accuracy of ADC with exemplar-based NME where the exemplars are used to estimate the old class prototype positions in the new feature space. We show in~\cref{fig:mem} that ADC outperforms NME using 20 exemplars per class for CIFAR-100 (total memory size of 2000 samples) and using 50 exemplars per class (total memory size of 10k samples) for TinyImageNet.

\subsection{Computational overhead of ADC}

\noindent Using ADC requires some additional computation to be made in-between each training session. In this section, we provide an estimation of the additional computation required by our method and compare it to the training time of a single task. At the end of each task, our method requires estimating the drift of each stored prototype (1 per old class) and for each of these, compute several adversarial samples starting from available current task samples. As a consequence, the training time of our method scales linearly with the number of classes. For each class, we compute 100 adversarial samples in a single batch and perform 3 training iterations. In order to perform one iteration, we need to compute the gradient of the adversarial loss with respect to the input image, whose cost is equivalent to the one of a normal training backward pass \cite{shafahi2019adversarial}. So, if we denote the number of classes by $N_c$, and the number of iterations by $N_i$, we need to perform $N_c \times N_i$ backward passes. In the case of CIFAR-100 and ImageNet-Subset divided in 10 tasks each containing 10 classes, this means an overhead of, $\sum_{t=1}^{9}{10 \times t \times 3} = 1350$ backward passes. In contrast, one new task is trained for 100 epochs with a batch size of 128 (39 batches per epochs with 10 tasks on CIFAR-100), which amounts to 3900 backward passes per task, and two times more for the first task (trained for 200 epochs). In total, our method increases the computational cost by 3.1\% on this setting. For the 5-task setting of CIFAR-100 and ImageNet-Subset, it increases by 2.5\%.

\subsection{Ablation Studies}
\begin{table}[t]
  \label{tab:ab}
  \centering
  \begin{minipage}{\linewidth}
    \begin{subtable}{\linewidth}
      \centering\small
      \caption{Impact of iterations ($\alpha=25$)}
      \label{tab:ab-iter}
      \begin{tabular}{cccccc}
        \toprule
        Iterations & 1 & 2 & 3 & 5 & 10 \\
        \midrule
        $A_{inc}$ & 60.94 & 61.23 & \textbf{61.35} & 61.25 & 60.93 \\
        $A_{last}$ & 45.96 & 46.45 & \textbf{46.48} & 45.95 & 45.28\\
        \bottomrule
      \end{tabular}
    \end{subtable}\vspace{2pt}
    
    \begin{subtable}{\linewidth}
      \centering\small
      \caption{Impact of $\alpha$ (iterations=3)}\label{tab:ab-alp}
      \begin{tabular}{cccccc}
        \toprule
        $\alpha$ & 1 & 10 & 25 & 50 & 100 \\
        \midrule
        $A_{inc}$ & 60.46 & 61.14 & \textbf{61.35} & 60.83 & 60.93 \\
       $ A_{last}$ & 44.89 & 46.43 & \textbf{46.48} & 45.19 & 45.28\\
        \bottomrule
        \end{tabular}
      \end{subtable}\vspace{2pt}%
      
      \begin{subtable}{\linewidth}
        \centering\small
        \caption{Impact of the number of samples}\label{tab:ab-sam}
        \begin{tabular}{ccccccc}
          \toprule
          Samples & 25 & 50 & 100 & 300 & 500 & 1000 \\
          \midrule
          $A_{inc}$ & 60.19 & 60.89 & 61.35 & 61.54 & \textbf{61.64} & 61.47 \\
          $A_{last}$ & 43.98 & 45.63 & 46.48 & 46.66 & \textbf{46.83} & 46.32\\
          \bottomrule
        \end{tabular}
      \end{subtable}
  \end{minipage}
  \caption{Impact of hyperparameters on CIFAR100 (T=10) setting using the proposed ADC method.}
  \vspace{-12pt}
\end{table}

In~\cref{tab:ab}, we conduct an analysis on the impact of various hyperparameters, including the number of iterations, $\alpha$, and the number of closest samples used for ADC, on CIFAR-100 (T=10) setting. Based on the observations in~\cref{tab:ab-iter}, we find that choosing even a very low number of iterations, specifically 3, yields favorable results when generating the perturbed images. Additionally, using $\alpha = 25$ achieves good accuracy for both incremental and final task evaluations in~\cref{tab:ab-alp}. Regarding the number of closest samples to the target old prototype,~\cref{tab:ab-sam} shows that the accuracy improves marginally on considering more than 100 samples. So, we take the 100 closest samples for our experiments which is computationally cheaper and yet achieves very good accuracy. Interestingly, even taking only the closest 25 samples achieves 2.62\% better accuracy than the runner-up method SDC.

\begin{figure}[t]
\centering
\includegraphics[width=1.0\linewidth]{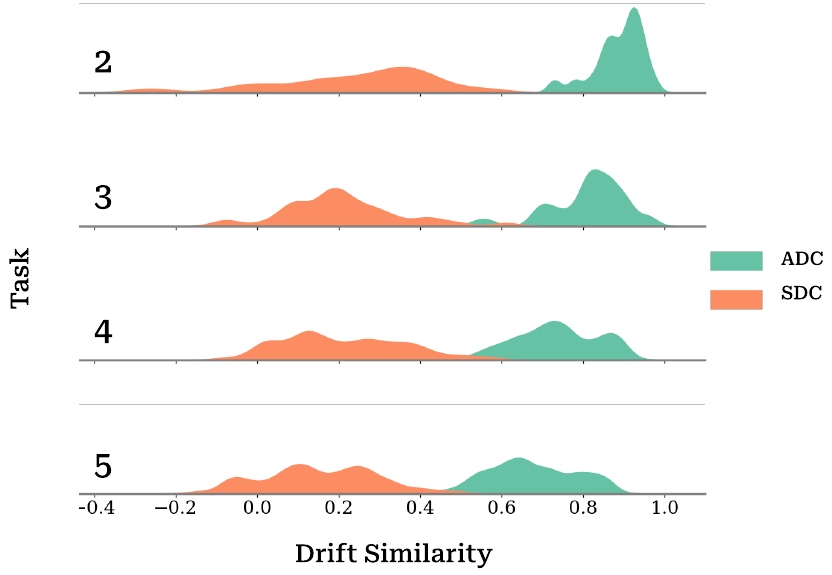}
\caption{Comparison between the two drift estimation methods SDC~\cite{Yu_2020_CVPR}, and the proposed ADC, on CIFAR-100 (5 tasks). We compute the drift for each class with the two methods and report the distribution of drift estimation quality, measured by computing the cosine similarity between the estimated drift vector and the true drift (obtained using old data), for all previous class prototypes.}
\label{fig:drift}
\vspace{-12pt}
\end{figure}

\noindent \textbf{Drift estimation quality:} We validate through~\cref{table1} and~\cref{table2} that the designed ADC method is giving better accuracy results than the previous SDC method for all datasets. As an additional verification, we check that this method was indeed better than SDC at estimating the old prototypes drift. To do so, we use both SDC and ADC on the same trained checkpoints on CIFAR-100 5-task settings and compare the estimated drift to the true drift computed using old data. We report the results in~\cref{fig:drift}, where we show the distribution of the estimated drift qualities. One drift per class is estimated and we compute the cosine similarity of estimated drift to the true drift. We see that for all training tasks, the drifts estimated with ADC are of better quality than the ones estimated with SDC. We observe that some class drift estimations with SDC have negative cosine similarity with the true drift. However, we also see that the estimation quality decreases slightly for later training tasks. Indeed, as the backbone drifts more and more, it gets harder to estimate the actual drift. The fact that we see this decrease more prominently for ADC might be because the similarities obtained by SDC are already centered around a low-value (0.15) after the second task, whereas the better ADC drift estimation is centered first around 0.9, to then decrease and reach a minimum average of 0.7. This validates that ADC is able to track the movement of the prototypes in the feature space.

\section{Conclusions}
In this study, we explored a drift compensation method for exemplar-free continual learning. Drawing inspiration from adversarial attack techniques, we introduced a novel approach called Adversarial Drift Compensation. This method involves generating samples from the new task data in a manner that adversarial images result in embeddings close to the old prototypes. This approach allows us to more accurately estimate the drift of old prototypes in class-incremental learning without the need for any exemplars.
Furthermore, we conducted an analysis of continual adversarial transferability, revealing an intriguing observation: generated samples for the old feature space (previous task) continue to behave similarly in the new feature space (current task). This sheds light on why the Adversarial Drift Compensation method performs exceptionally well.
Through a series of experiments, we demonstrated that ADC effectively tracks the drift of class distributions in the embedding space, surpassing existing exemplar-free class-incremental learning methods on several standard benchmarks. Importantly, these improvements are achieved without imposing extensive computational overhead or requiring a large memory footprint.

\noindent\textbf{Limitations.} The ADC method, as currently designed, requires the access to the task boundaries during training in order to trigger the computation of the old prototypes drift and to access a big enough quantity of current data. The method would for instance be more challenging to use and would require changes in order to be applied in the online continual learning setting, or the continual few-shot learning setting where only a small amount of current data is available. Future work can explore these directions.

\paragraph{Acknowledgement.} We acknowledge projects TED2021-132513B-I00 and PID2022-143257NB-I00, financed by MCIN/AEI/10.13039/501100011033 and FSE+ and the Generalitat de Catalunya CERCA Program.
This work was partially funded by the European Union under the Horizon Europe Program (HORIZON-CL4-2022-HUMAN-02) under the project ``ELIAS: European Lighthouse of AI for Sustainability", GA no. 101120237. Bartłomiej Twardowski acknowledges the grant RYC2021-032765-I.


\clearpage
\setcounter{page}{1}
\maketitlesupplementary


\section{Training settings and hyperparameters}
\label{sec:settings}

Since the current approaches are not designed and optimized for the \textit{small-start} settings used in our work, we adapt relevant methods and optimize them for these settings and achieve comparable baselines to our approach.
We list the exact experimental settings to enable reproducibility.

\noindent \textbf{Augmentations:} As implemented in PyCIL~\cite{zhou2023pycil}, for CIFAR-100, we use the same augmentation policy which consists of small random transformations like contrast or brightness changes. Similarly, for the other datasets, we use the default set of augmentations which include random crop and random horizontal flip. For a fair evaluation, we use the same set augmentations for all the methods.

\noindent \textbf{LwF:} 
In CIFAR-100, TinyImageNet and ImageNet-Subset datasets for the first task, similar to PyCIL~\cite{zhou2023pycil}, we use a starting learning rate of 0.1, momentum of 0.9, batch size of 128, weight decay of 5e-4 and trained for 200 epochs, with the learning rate reduced by a factor of 10 after 60, 120, and 160 epochs, respectively. For subsequent tasks, we used an initial learning rate of 0.05 for CIFAR-100 and ImageNet-Subset and 0.001 for TinyImageNet. The learning rate is reduced by a factor of 10 after 45 and 90 epochs and trained for a total of 100 epochs. We set the the temperature to 2 and the regularization strength to 10 for CIFAR-100 and TinyImageNet and 5 for ImageNet-Subset. 
For the fine-grained datasets, we use a learning rate of 0.01 for the first task and a learning rate of 0.005 for subsequent tasks. The regularization strength is set to 20.

For the NCM classifier, SDC and our proposed method ADC, we use the same training settings as LwF.

\noindent \textbf{SDC:} For SDC~\cite{Yu_2020_CVPR}, we set the hyperparameters $\sigma=0.3$ for CIFAR-100, TinyImagenet and for the fine-grained datasets. For Imagenet-Subset, we set $\sigma=1.0$.

\noindent \textbf{PASS:} We follow the implementation of PASS~\cite{zhu2021prototype} from PyCIL~\cite{zhou2023pycil} and set $\lambda_{fkd} = 10$ and $\lambda_{proto} = 10$.

\noindent \textbf{SSRE:} We follow the implementation of SSRE~\cite{zhu2022self} from PyCIL~\cite{zhou2023pycil} and set $\lambda_{fkd} = 10$ and $\lambda_{proto} = 10$.

\noindent \textbf{FeTrIL:} For first task, we use a learning rate of 0.1 for CIFAR-100, TinyImageNet and ImageNet-Subset and follow the exact same settings as the original implementation~\cite{petit2023fetril}. For the fine-grained datasets, we use a learning rate of 0.01 for the first task. 

\noindent \textbf{FeCAM:} We use the same training setting as LwF for the first task training. FeCAM~\cite{goswami2023fecam} requires no training after the first task and stores the prototypes and covariance matrices from all the classes. Similar to the original implementation, we use the covariance shrinkage hyperparameters of (1,1) and the Tukey's normalization value of 0.5.

\noindent \textbf{ADC:} We use a $\alpha$ value of 25, iterations $i = 3$ and number of closest samples $m=100$ for all the datasets.

\begin{figure}[t]
\centering
\includegraphics[width=\linewidth]{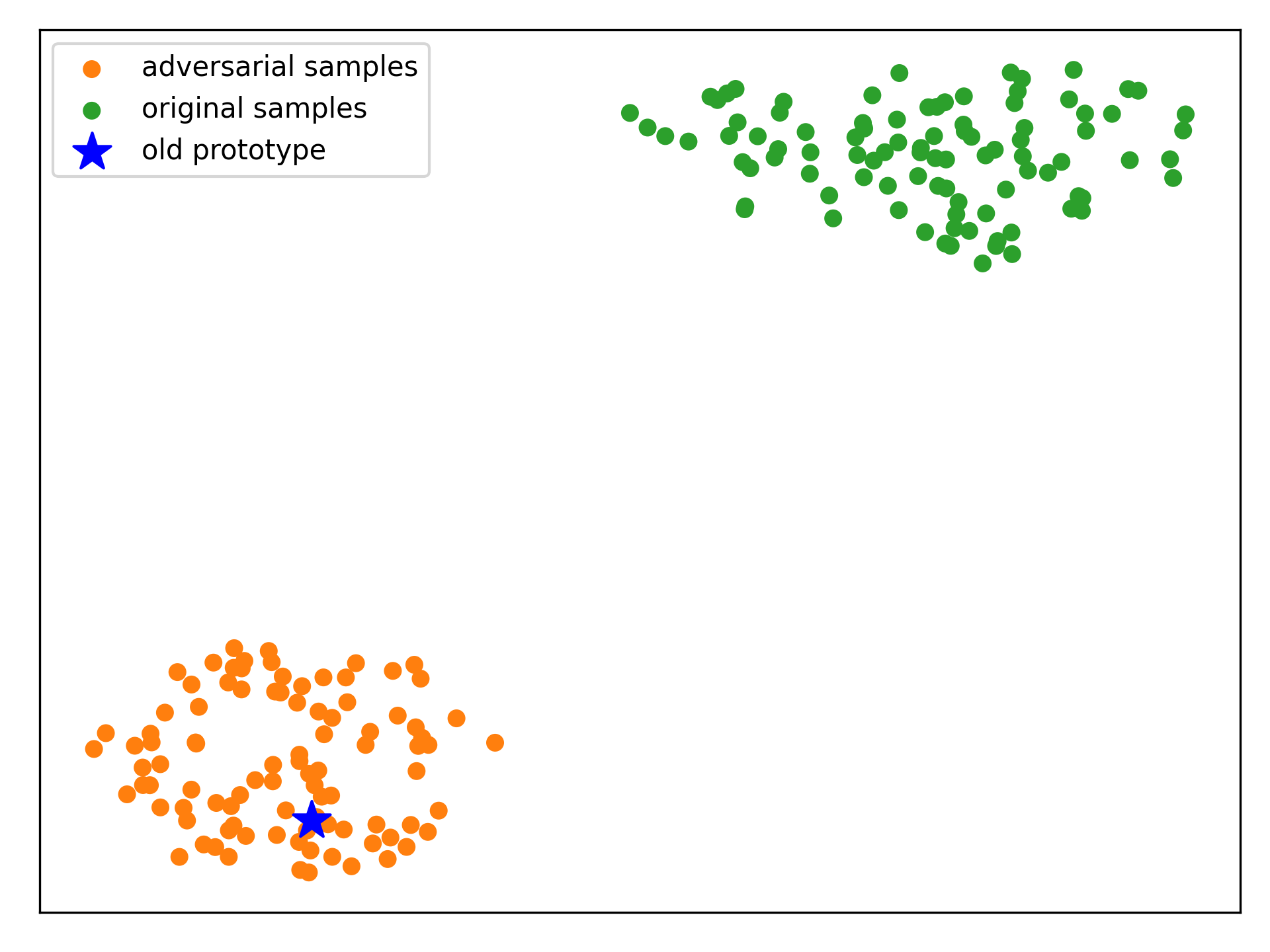}
\caption{The t-SNE plot demonstrates that the adversarial samples generated using our proposed method lie close to the target old class prototype compared to the closest current task samples (in green) and can thus be reliably used for drift estimation.} 
\label{fig:t-sne}
\vspace{-12pt}
\end{figure}


\begin{table*}[t]
\begin{center}
\resizebox{0.7\textwidth}{!}{
\begin{tabular}{lcccc}
\toprule
    \multirow{2}{*}{Method}        & \multicolumn{2}{c}{T = 5} & \multicolumn{2}{c}{T=10} \\
           & $A_{last}$     & $A_{inc}$    & $A_{last}$    & $A_{inc}$      \\
           \midrule
LwF~\cite{li2017learning}         &      $45.67 \pm 1.37$        &     $62.12 \pm 1.08$      &      $26.59 \pm 2.44$       &    $45.60 \pm 2.52$      \\
NCM        &      \underline{$52.68 \pm 0.57$}        &     \underline{$65.91 \pm 0.4$}       &      $40.53 \pm 2.74$       &     \underline{$56.21 \pm 3.55$}         \\
SDC~\cite{Yu_2020_CVPR}        &      $52.26 \pm 3.0$        &      $63.88 \pm 1.23$      &      \underline{$40.65 \pm 1.82$}       &     $56.18 \pm 3.0$       \\
FeTrIL~\cite{petit2023fetril}     &      $45.52 \pm 0.33$        &      $60.84 \pm 0.46$      &      $37.0 \pm 0.57$       &     $52.08 \pm 0.51$           \\
FeCAM~\cite{goswami2023fecam}      &   $46.93 \pm 0.34$  &  $61.49 \pm 0.55$  &  $33.13 \pm 0.93$  &  $48.10 \pm 1.27$          \\
\midrule
ADC (Ours) &      $\mathbf{58.12 \pm 1.42}$        &       $\mathbf{69.29 \pm 1.17}$     &      $\mathbf{45.43 \pm 3.03}$       &      $\mathbf{59.59 \pm 4.11}$         \\
\bottomrule
\end{tabular}
}
\caption{Evaluation of EFCIL methods with mean and standard deviation using 5 random seeds for CIFAR-100. Best results in \textbf{bold} and second best results are \underline{underlined}.}
\label{random_table1}
\end{center}
\vspace{-15pt}
\end{table*}

\begin{table*}[t]
\begin{center}
\resizebox{0.7\textwidth}{!}{
\begin{tabular}{lcccc}
\toprule
    \multirow{2}{*}{Method}        & \multicolumn{2}{c}{T = 5} & \multicolumn{2}{c}{T=10} \\
           & $A_{last}$     & $A_{inc}$    & $A_{last}$    & $A_{inc}$      \\
           \midrule
LwF~\cite{li2017learning}         &        $39.03 \pm 0.43$       &     $50.96 \pm 0.93$       &      $27.75 \pm 0.51$       &     $40.04 \pm 0.96$   \\
NCM        &           $38.76 \pm 0.18$        &      $51.74 \pm 0.8$      &      $28.07 \pm 0.97$        &     \underline{$42.86 \pm 0.98$}     \\
SDC~\cite{Yu_2020_CVPR}        &      \underline{$40.28 \pm 0.37$}        &    \underline{$52.21 \pm 0.89$} &  \underline{$28.15 \pm 0.67$}       &       $42.09 \pm 1.03$       \\
FeTrIL~\cite{petit2023fetril}     &       $29.94 \pm 0.83$       &     $45.08 \pm 0.98$       &      $23.6 \pm 0.42$        &     $37.41 \pm 0.63$       \\
FeCAM~\cite{goswami2023fecam}      &  $26.03 \pm 0.49$  &  $41.11 \pm 0.93$  &  $23.78 \pm 0.48$  &  $37.30 \pm 1.23$           \\
\midrule
ADC (Ours) &       $\mathbf{41.29 \pm 0.47}$        &     $\mathbf{52.36 \pm 0.95}$       &     $\mathbf{32.68 \pm 0.43}$         &    $\mathbf{44.89 \pm 1.03}$       \\
\bottomrule
\end{tabular}
}
\caption{Evaluation of EFCIL methods with mean and standard deviation using 5 random seeds for TinyImageNet. Best results in \textbf{bold} and second best results are \underline{underlined}.}
\label{random_table2}
\end{center}
\vspace{-15pt}
\end{table*}

\begin{table*}[h!]
\begin{center}
\resizebox{0.7\textwidth}{!}{
\begin{tabular}{lcccc}
\toprule
    \multirow{2}{*}{Method}       & \multicolumn{2}{c}{T = 5} & \multicolumn{2}{c}{T=10} \\
           & $A_{last}$     & $A_{inc}$    & $A_{last}$    & $A_{inc}$      \\
           \midrule
LwF~\cite{li2017learning}         &      $49.16 \pm 1.34$        &     $68.88 \pm 0.74$      &      $34.18 \pm 3.69$       &    $57.96 \pm 3.64$      \\
NCM        &     $56.80 \pm 1.90$        &     $72.45 \pm 0.87$       &      \underline{$44.04 \pm 1.93$}       &     $64.43 \pm 0.43$         \\
SDC~\cite{Yu_2020_CVPR}        &      \underline{$59.62 \pm 1.56$}        &      \underline{$74.44 \pm 0.76$}      &      $42.68 \pm 1.88$       &     \underline{$65.26 \pm 0.59$}       \\
FeTrIL~\cite{petit2023fetril}     &      $50.52 \pm 1.06$        &      $65.64 \pm 0.95$      &      $40.74 \pm 0.50$       &     $56.34 \pm 0.76$           \\
FeCAM~\cite{goswami2023fecam}      &    $53.83 \pm 0.46$  &  $67.88 \pm 0.67$  &  $42.46 \pm 0.89$  &  $57.93 \pm 1.45$        \\
\midrule
ADC (Ours) &      $\mathbf{61.62 \pm 0.93}$        &    $\mathbf{75.29 \pm 0.61}$     &      $\mathbf{47.62 \pm 1.55}$  &  $\mathbf{67.03 \pm 0.46}$         \\
\bottomrule
\end{tabular}
}
\caption{Evaluation of EFCIL methods with mean and standard deviation using 5 random seeds for ImageNet-Subset. Best results in \textbf{bold} and second best results are \underline{underlined}.}
\label{random_table3}
\end{center}
\vspace{-20pt}
\end{table*}

\section{Robustness to different class orders}

In CIL, the order of classes can influence the performance and thus we shuffle the class orders and observe how ADC and the existing methods like LwF, NCM, SDC, FeTrIL and FeCAM perform. While we used the seed 1993 following previous works~\cite{petit2023fetril,masana2020class,rebuffi2017icarl,Yu_2020_CVPR} for the results reported in the main paper, here we use four different seeds 0, 1, 2, 3 and report the mean and standard deviation using these 5 seeds for both the last task accuracy $A_{last}$ and the average incremental accuracy $A_{inc}$ in~\cref{random_table1,random_table2,random_table3}.
The proposed method ADC outperforms SDC and NCM consistently across all settings on CIFAR-100, TinyImageNet and ImageNet-Subset. This demonstrates the robustness of ADC which improves over the existing methods irrespective of the class order.

\section{Perturbation guarantee} 
We specifically select the closest samples to each old prototype, one at a time (see Algorithm 1) to ensure we generate adversarial samples for all the old classes.
On CIFAR100 (T=10), we get an average of 59 samples out of 100, which are successfully perturbed for all old classes after the last task. While performing 5 iterations (instead of 3) generates an average of 69 successful perturbations for old classes, this does not lead to a significant accuracy change (Tab. 3a). Therefore, we have used 3 iterations in our implementation.

We analyze the position of the closest current task samples and the generated adversarial samples with respect to a target old class prototype in the old feature space using a t-SNE plot in~\cref{fig:t-sne}. We observe that the adversarial samples lie close to the prototype, while the original samples are distant from the prototype. This validates the effectiveness of the adversarial attack in the old feature space and shows how new samples obtained using targeted adversarial attacks can be used to represent old classes.
These adversarial samples behave as pseudo-exemplars and can now be used to estimate the drift of prototypes from the old to the new feature space.

\section{Prompt-based Methods}
Prompt-based methods~\cite{wang2022learning,wang2022dualprompt,smith2023coda} aim to learn prompt parameters that can be used with frozen pre-trained models without updating the parameters of the model. A recent work, HiDe-Prompt~\cite{wang2024hierarchical} also freezes the pre-trained ViT backbones and proposes an ensemble strategy for using prompts. 
Different from them, our objective is to continually learn new representations and update the backbone at every task. 
These methods have static features due to the frozen backbone and avoid the feature drift problem we are tackling. We think it is unfair to compare the performance of frozen pre-trained models with our method (training from scratch and updating the backbone).
While freezing pre-trained models works well for mainstream datasets, it is crucial to update the backbone and learn new representations 
for training domain-specific models for data that are not commonly seen in pre-trained data, and thus it is necessary to develop drift compensation methods. Janson et al.~\cite{janson2022simple} show that while pre-trained models with a simple NCM baseline work similar to L2P on Cifar100, they struggle on ImageNet-R with data of different styles like cartoon, graffiti, and origami.

\section{Visualization of adversarial images}
\begin{figure*}[t]
     \centering
     \begin{subfigure}[b]{0.23\textwidth}
         \centering
         \begin{subfigure}[b]{\textwidth}
             \centering
             \hspace{-0.45em}
             \includegraphics[width=\textwidth]{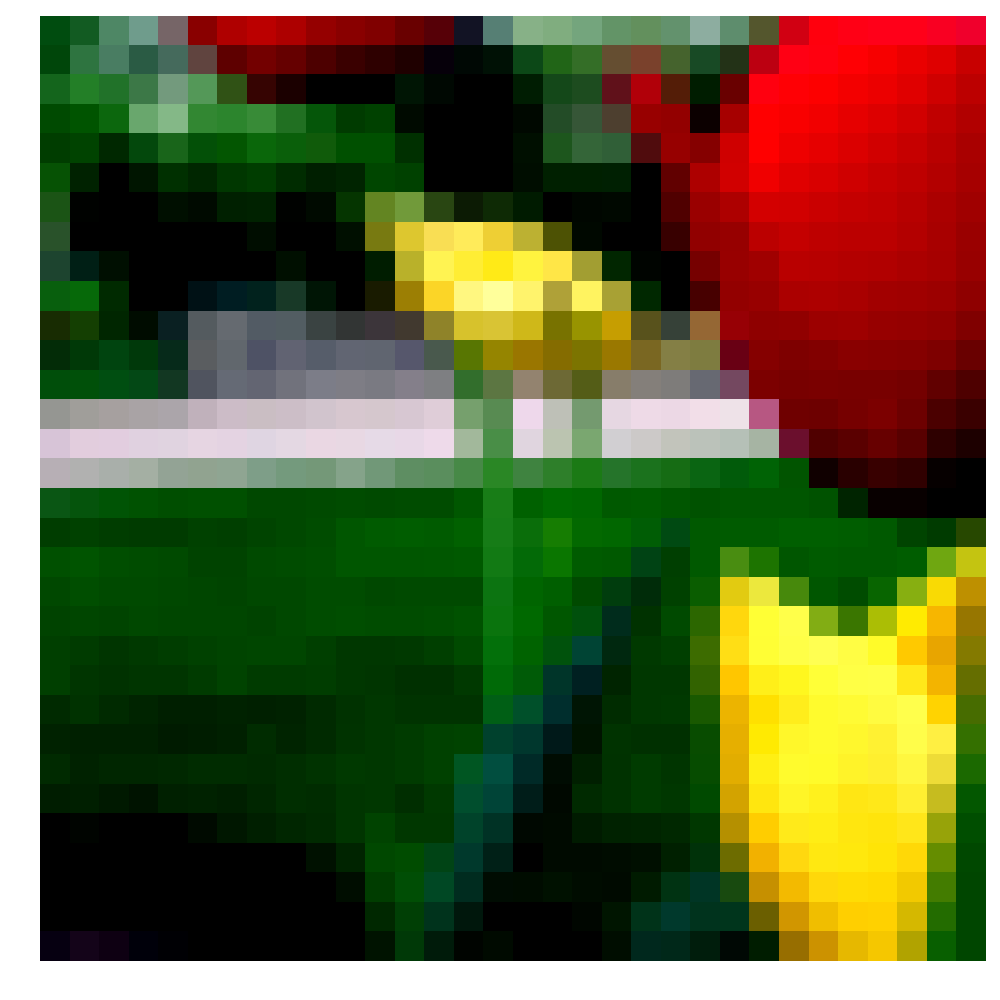}
             \vspace{-1.2em}
         \end{subfigure}

         \begin{subfigure}[b]{\textwidth}
             \centering
             \includegraphics[width=\textwidth]{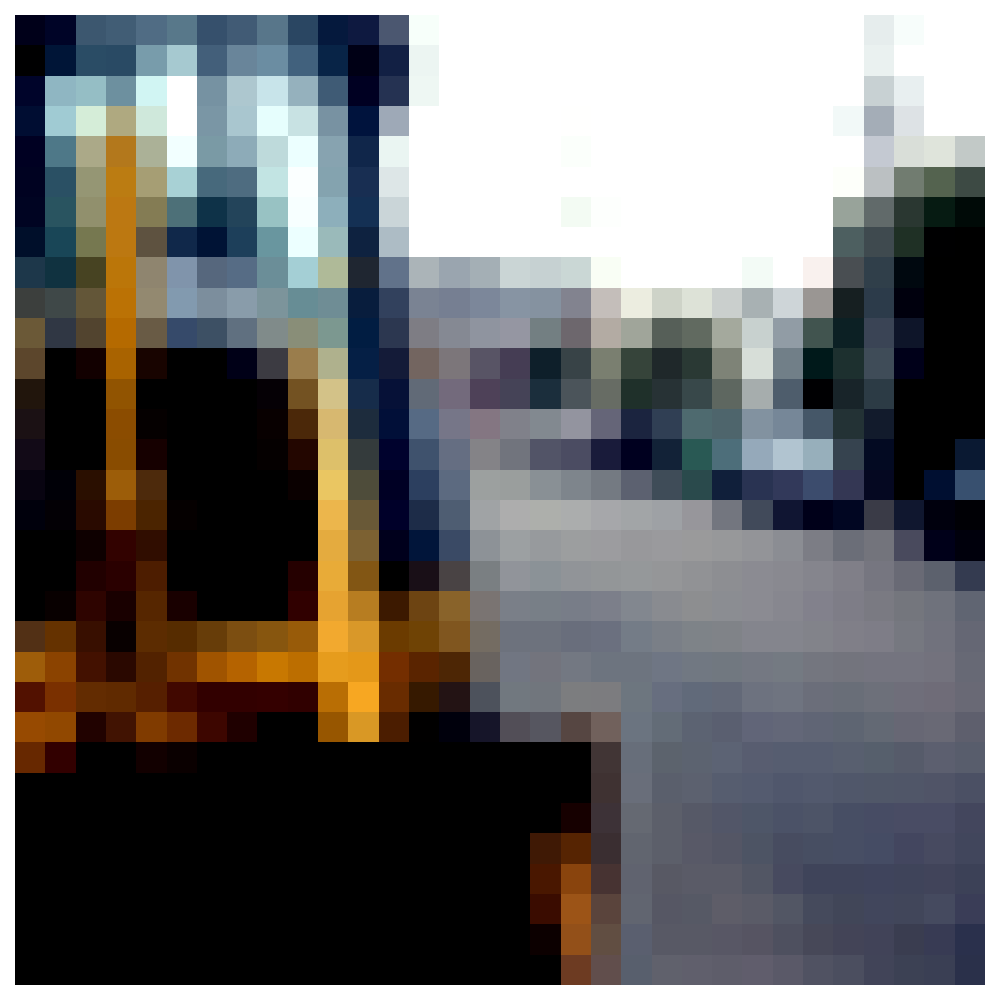}
             \vspace{-1em}
         \end{subfigure}

         \begin{subfigure}[b]{\textwidth}
             \centering
             \includegraphics[width=\textwidth]{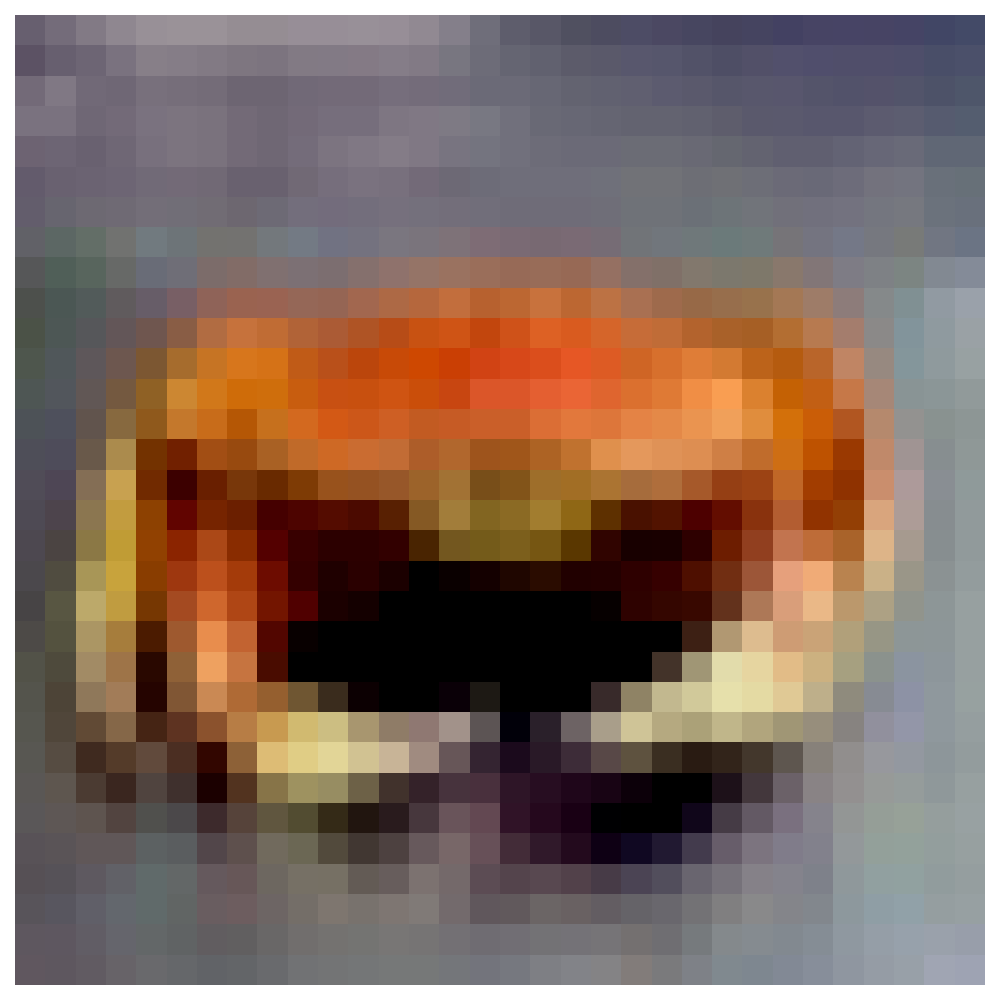}
             \vspace{-1em}
         \end{subfigure}
         
         \begin{subfigure}[b]{\textwidth}
             \centering
             \includegraphics[width=\textwidth]{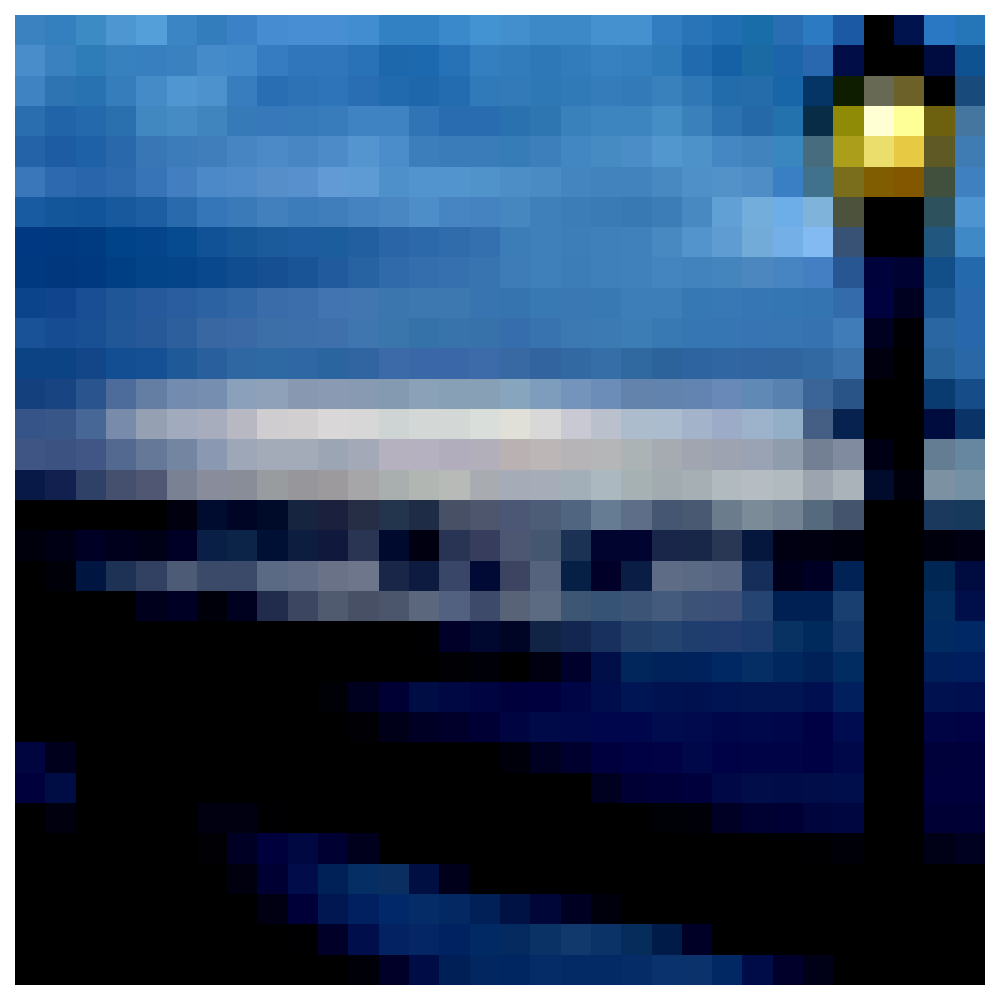}
             \vspace{-1em}
         \end{subfigure}
         
         \begin{subfigure}[b]{\textwidth}
             \centering
             \includegraphics[width=\textwidth]{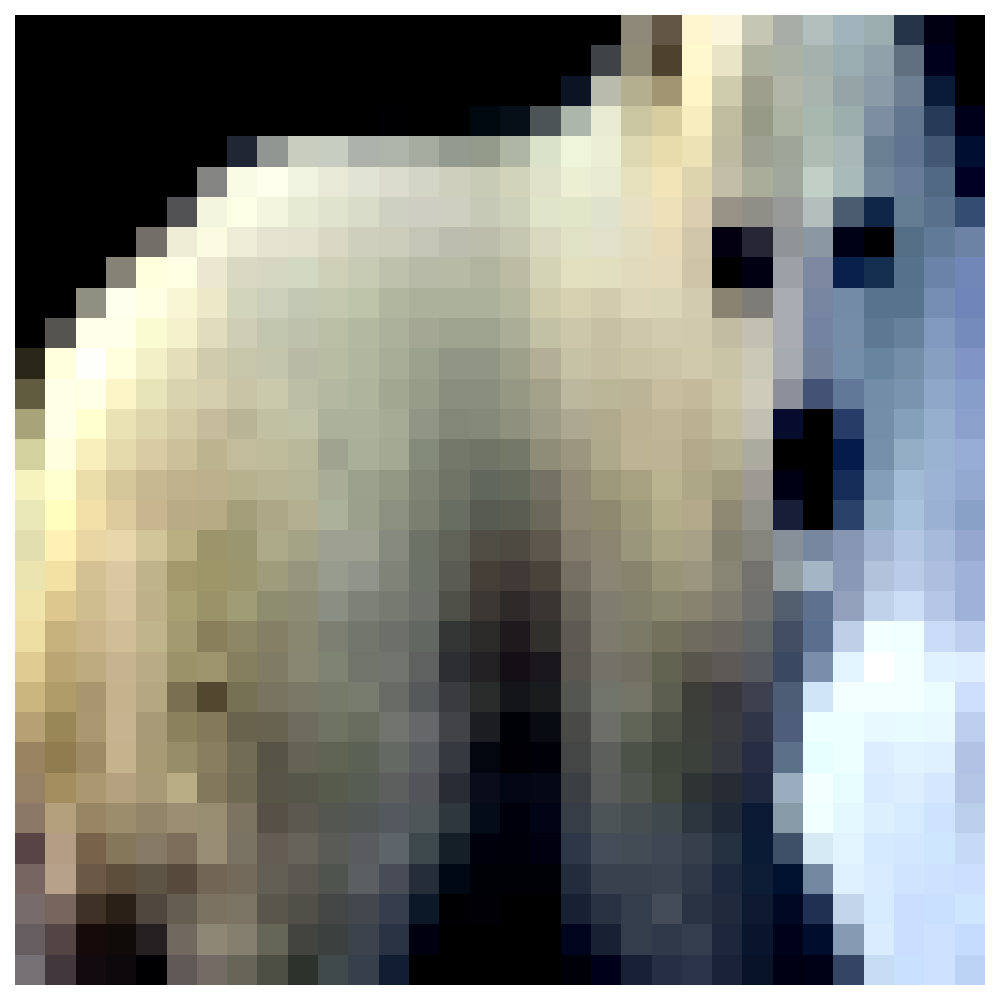}
             \vspace{-1em}
         \end{subfigure}
         \caption{Original image}
    \end{subfigure}
    \begin{subfigure}[b]{0.23\textwidth}
         \centering
         \vspace{-1em}
         \begin{subfigure}[b]{\textwidth}
             \centering
             \hspace{-0.45em}
             \includegraphics[width=\textwidth]{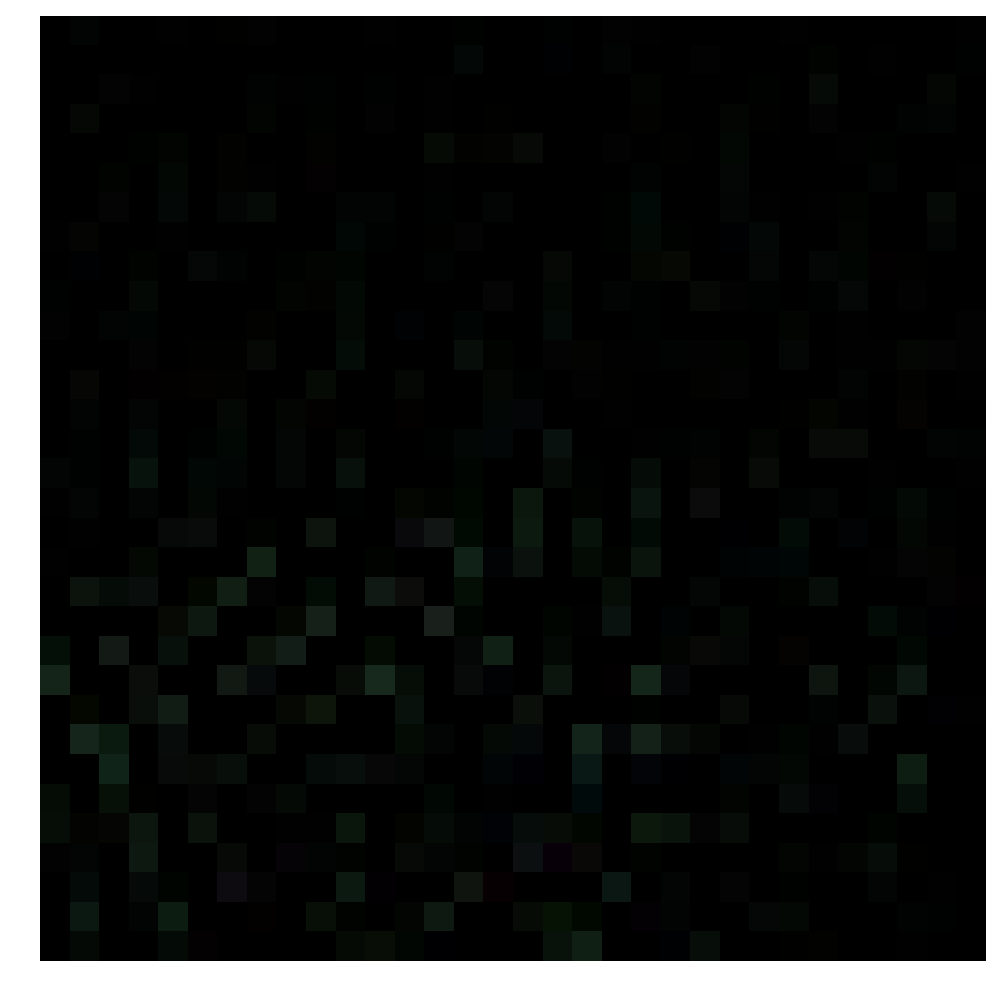}
             \vspace{-1.2em}
         \end{subfigure}

         \begin{subfigure}[b]{\textwidth}
             \centering
             \includegraphics[width=\textwidth]{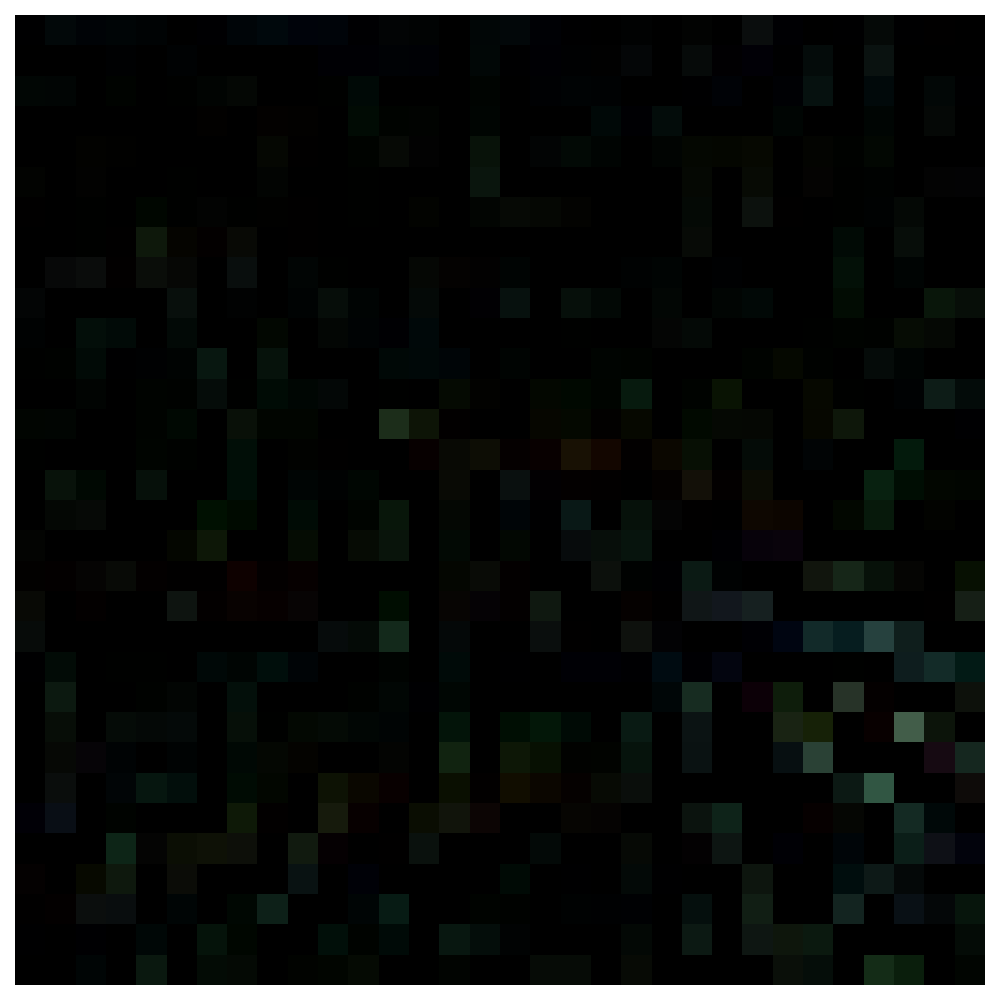}
             \vspace{-1em}
         \end{subfigure}

         \begin{subfigure}[b]{\textwidth}
             \centering
             \includegraphics[width=\textwidth]{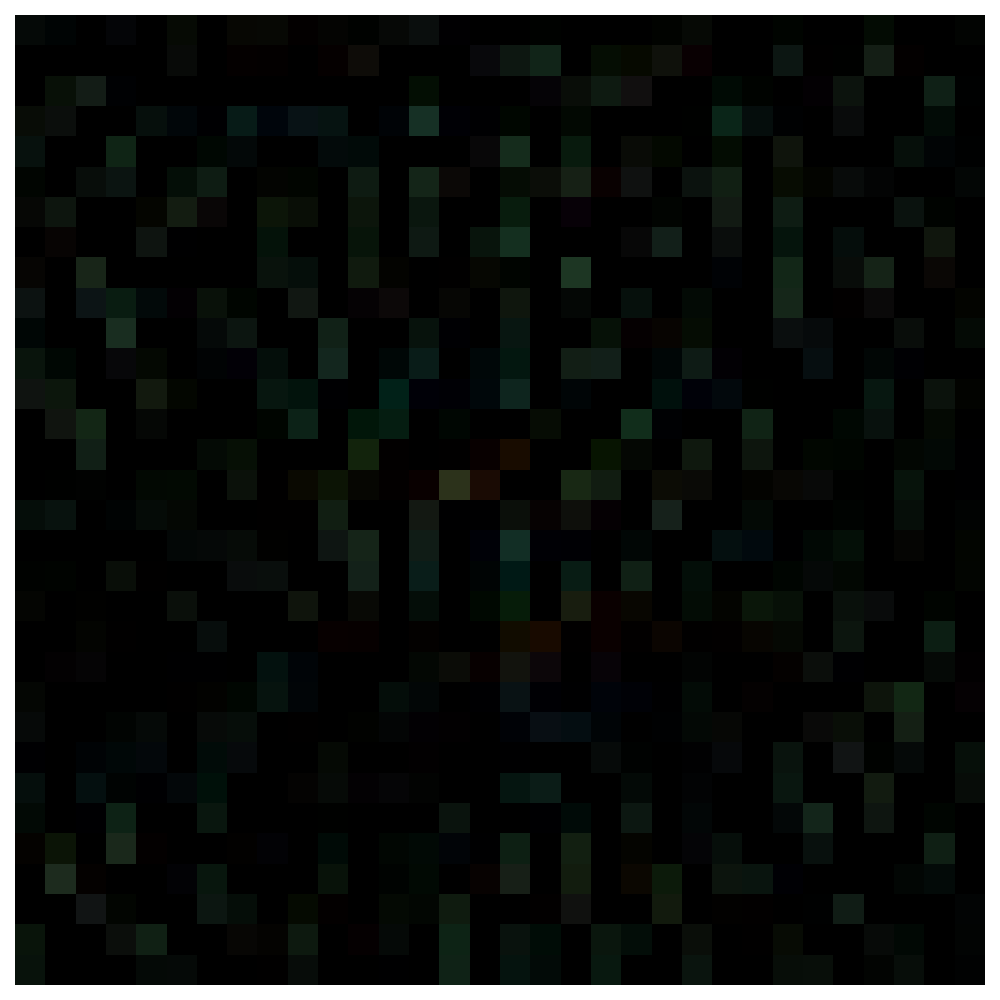}
             \vspace{-1em}
         \end{subfigure}

         \begin{subfigure}[b]{\textwidth}
             \centering
             \includegraphics[width=\textwidth]{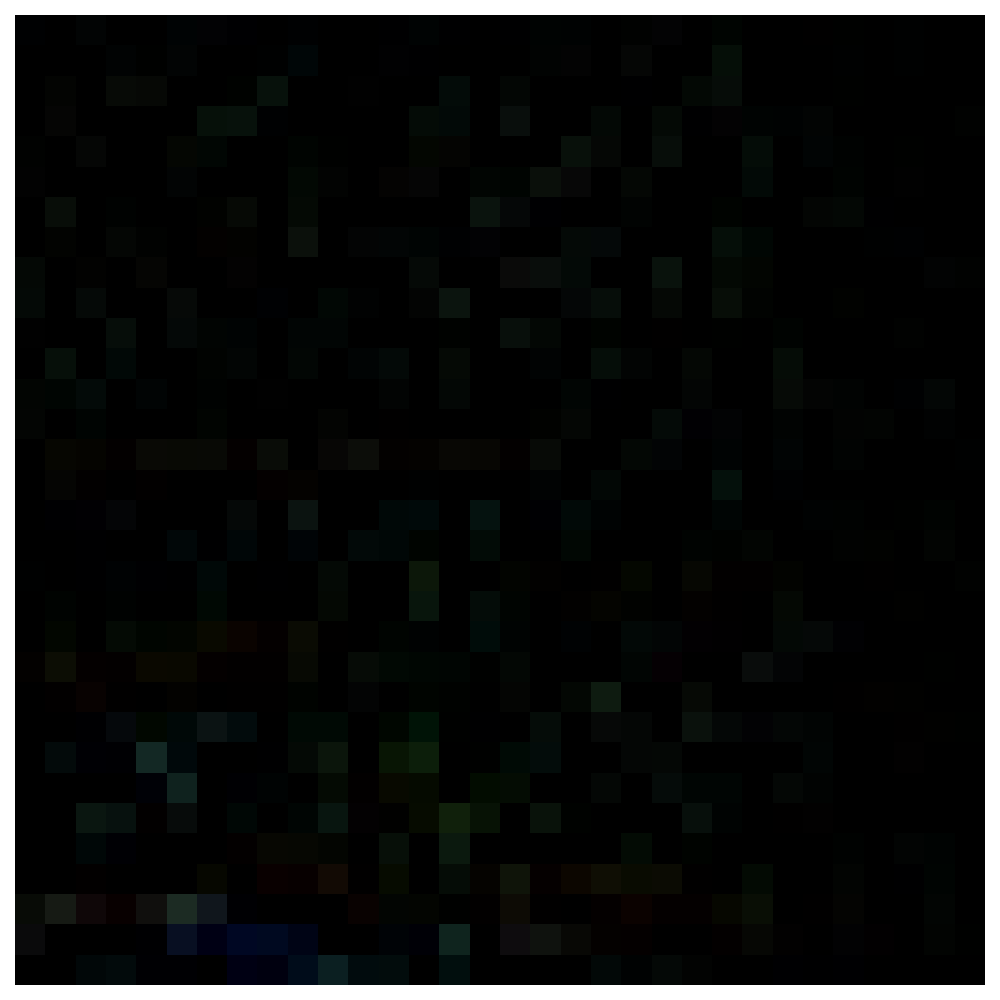}
             \vspace{-1em}
         \end{subfigure}

         \begin{subfigure}[b]{\textwidth}
             \centering
             \includegraphics[width=\textwidth]{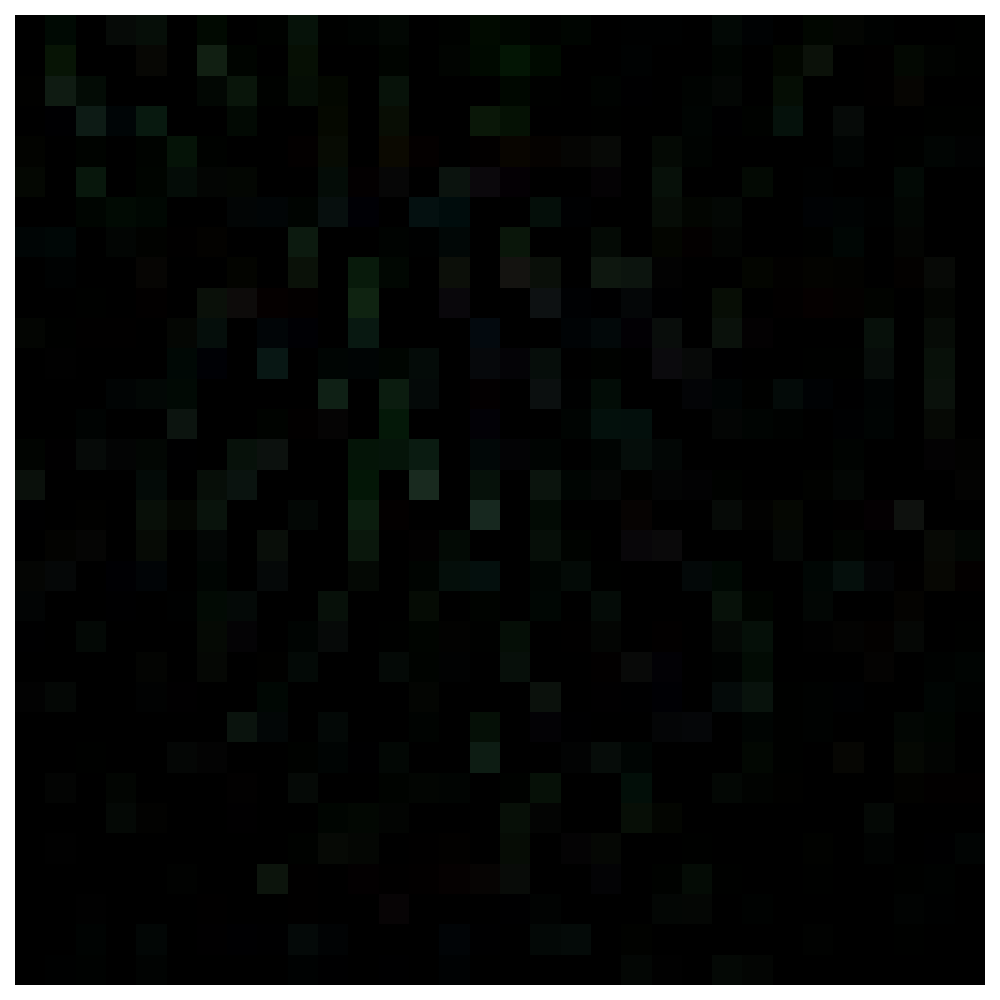}
             \vspace{-1em}
         \end{subfigure}

         \caption{Adversarial perturbation}
    \end{subfigure}
    \begin{subfigure}[b]{0.23\textwidth}
         \centering
         \vspace{-1em}
         \begin{subfigure}[b]{\textwidth}
             \centering
             \hspace{-0.45em}
             \includegraphics[width=\textwidth]{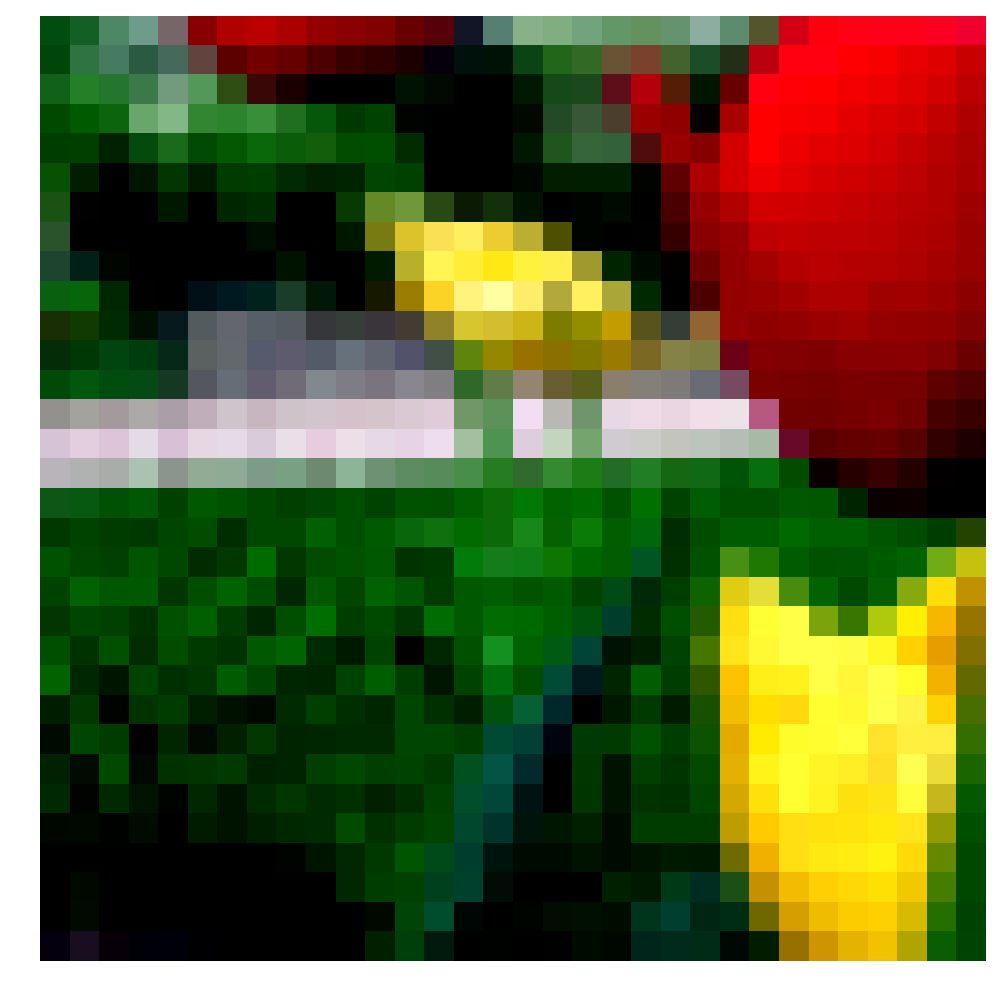}
             \vspace{-1.2em}
         \end{subfigure}

         \begin{subfigure}[b]{\textwidth}
             \centering
             \includegraphics[width=\textwidth]{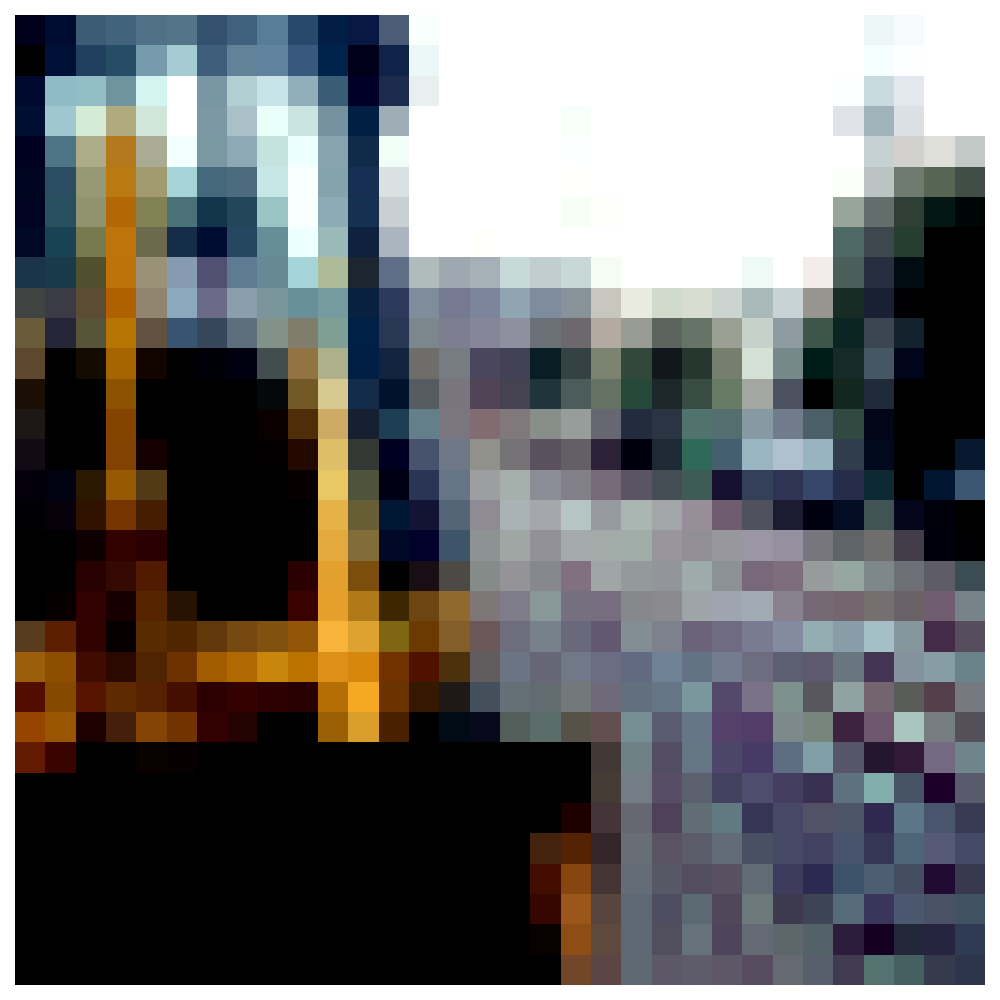}
             \vspace{-1em}
         \end{subfigure}

         \begin{subfigure}[b]{\textwidth}
             \centering
             \includegraphics[width=\textwidth]{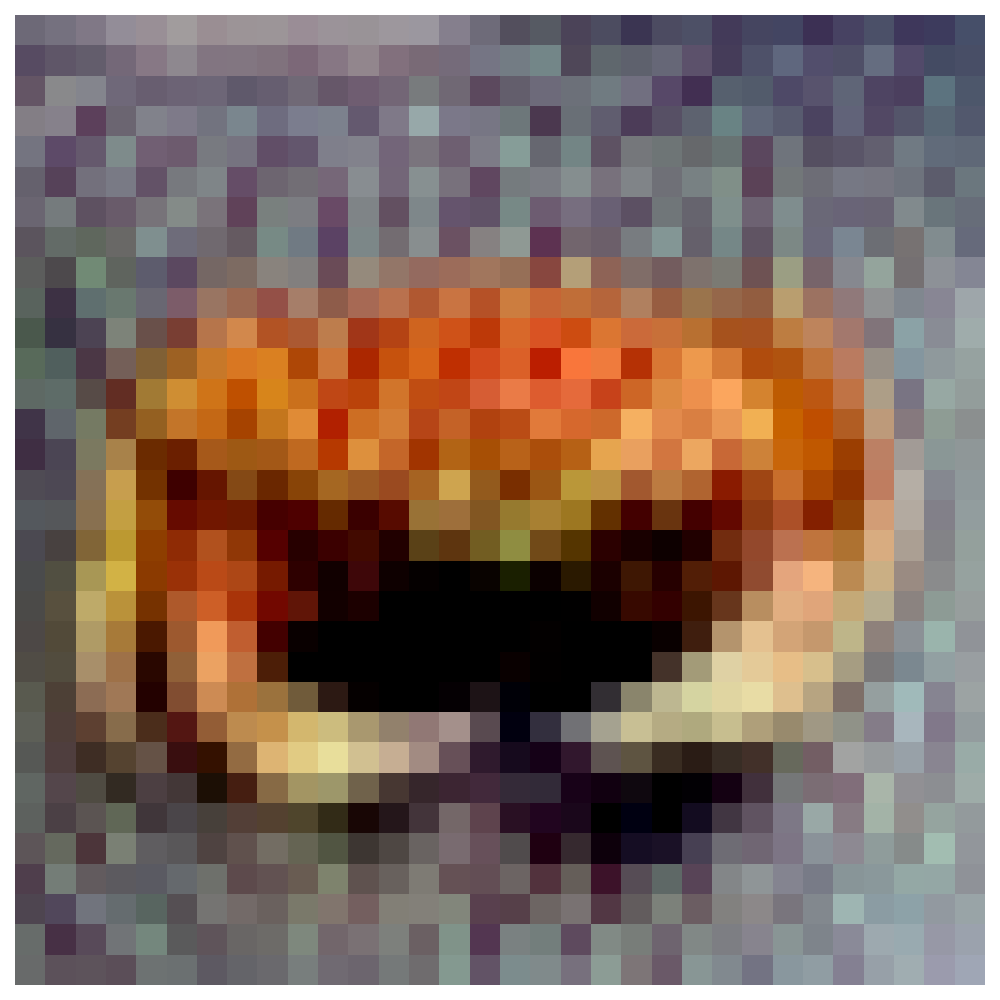}
             \vspace{-1em}
         \end{subfigure}

         \begin{subfigure}[b]{\textwidth}
             \centering
             \includegraphics[width=\textwidth]{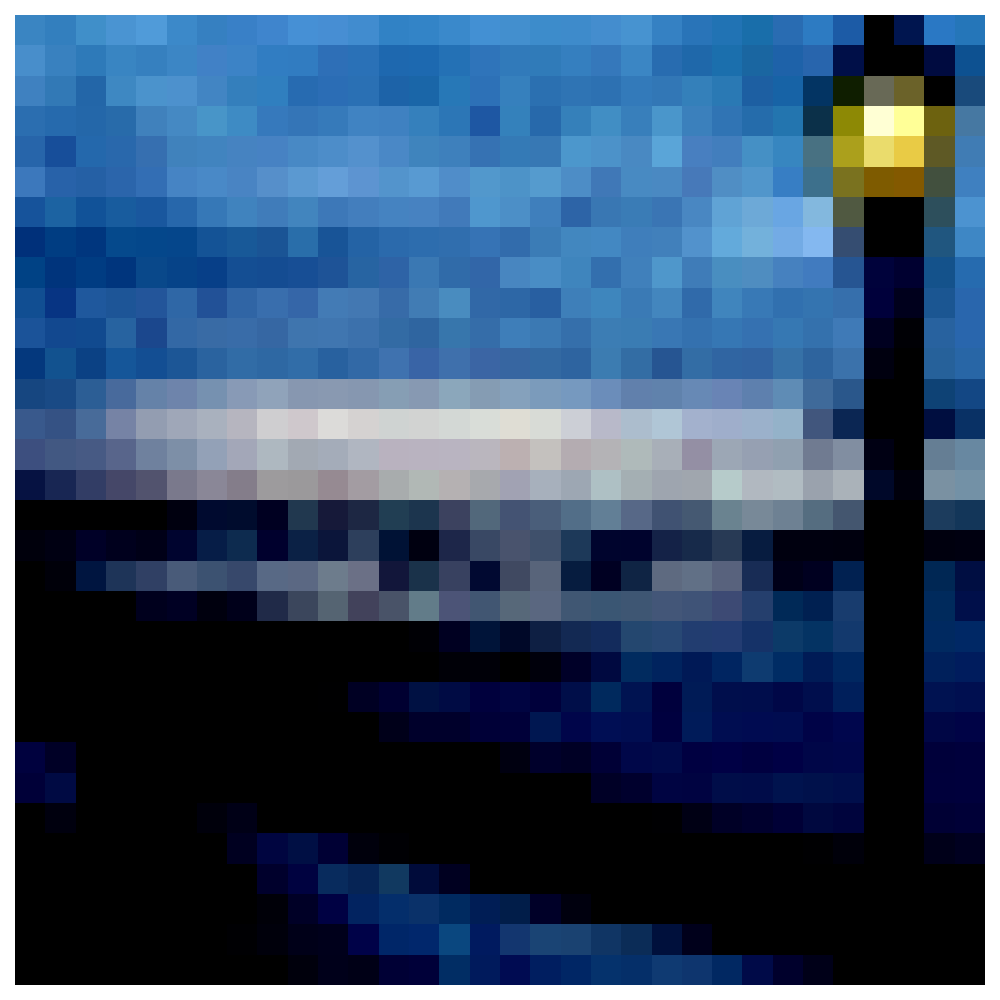}
             \vspace{-1em}
         \end{subfigure}

         \begin{subfigure}[b]{\textwidth}
             \centering
             \includegraphics[width=\textwidth]{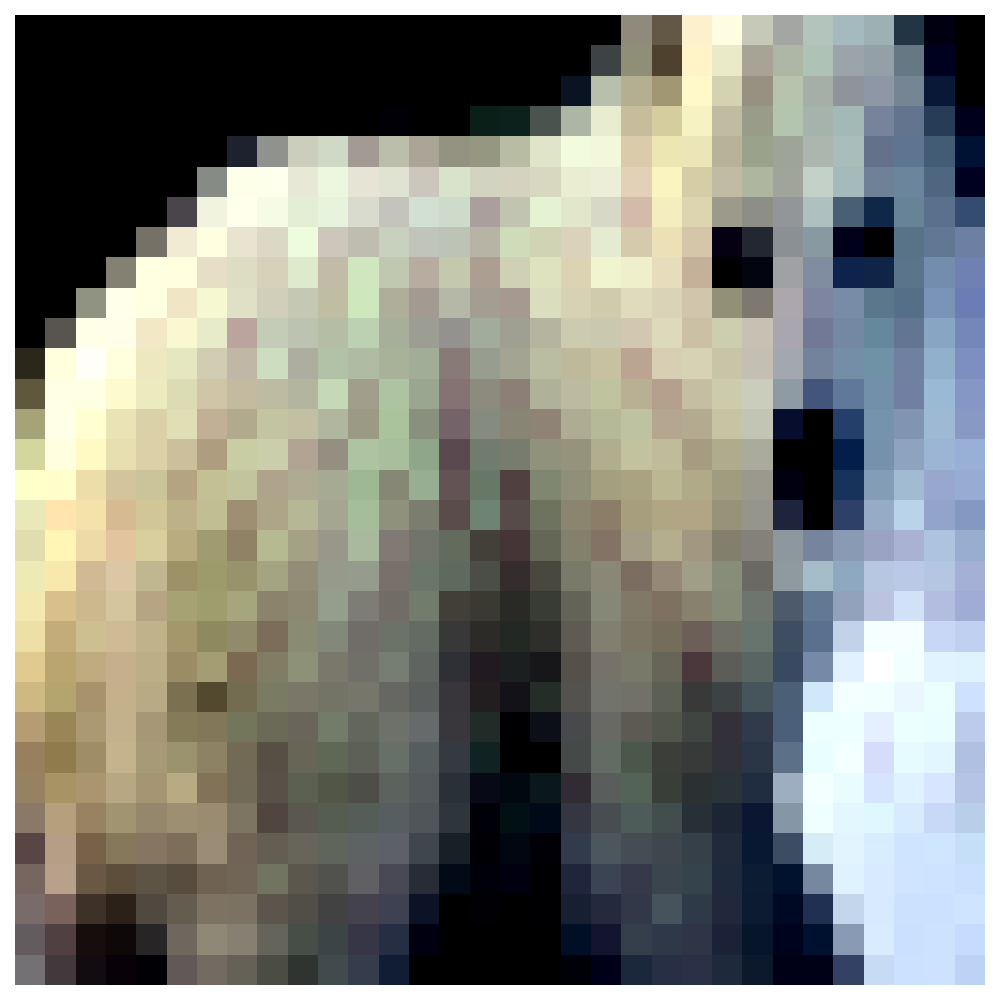}
             \vspace{-1em}
     \end{subfigure}
     \caption{Adversarial image}
    \end{subfigure}

    \caption{Visualization of image, perturbation and the corresponding adversarial image for some samples from CIFAR-100.}
    \label{fig:cifar_combo}
\end{figure*}

\begin{figure*}[t]
     \centering
     \begin{subfigure}[b]{0.23\textwidth}
         \centering
         \begin{subfigure}[b]{\textwidth}
             \centering
             \hspace{-0.45em}
             \includegraphics[width=\textwidth]{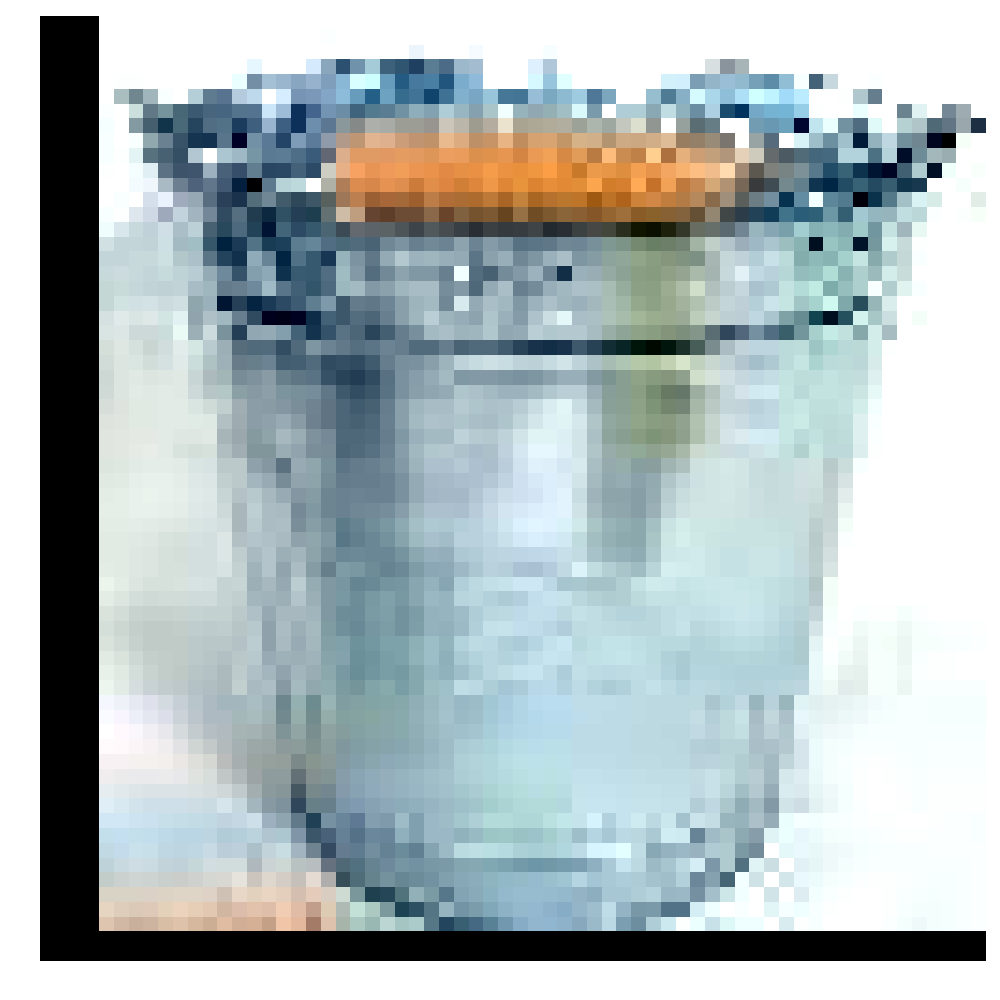}
             \vspace{-1.2em}
         \end{subfigure}

         \begin{subfigure}[b]{\textwidth}
             \centering
             \includegraphics[width=\textwidth]{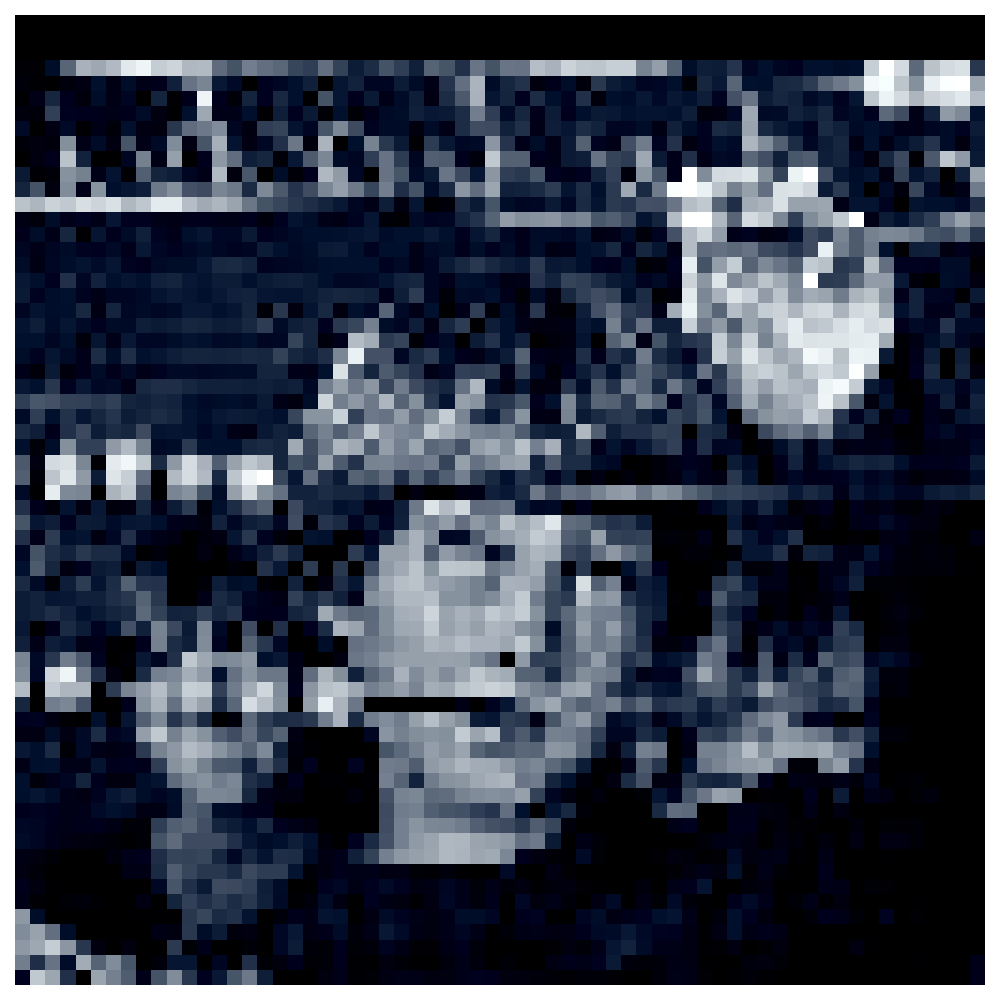}
             \vspace{-1em}
         \end{subfigure}

         \begin{subfigure}[b]{\textwidth}
             \centering
             \includegraphics[width=\textwidth]{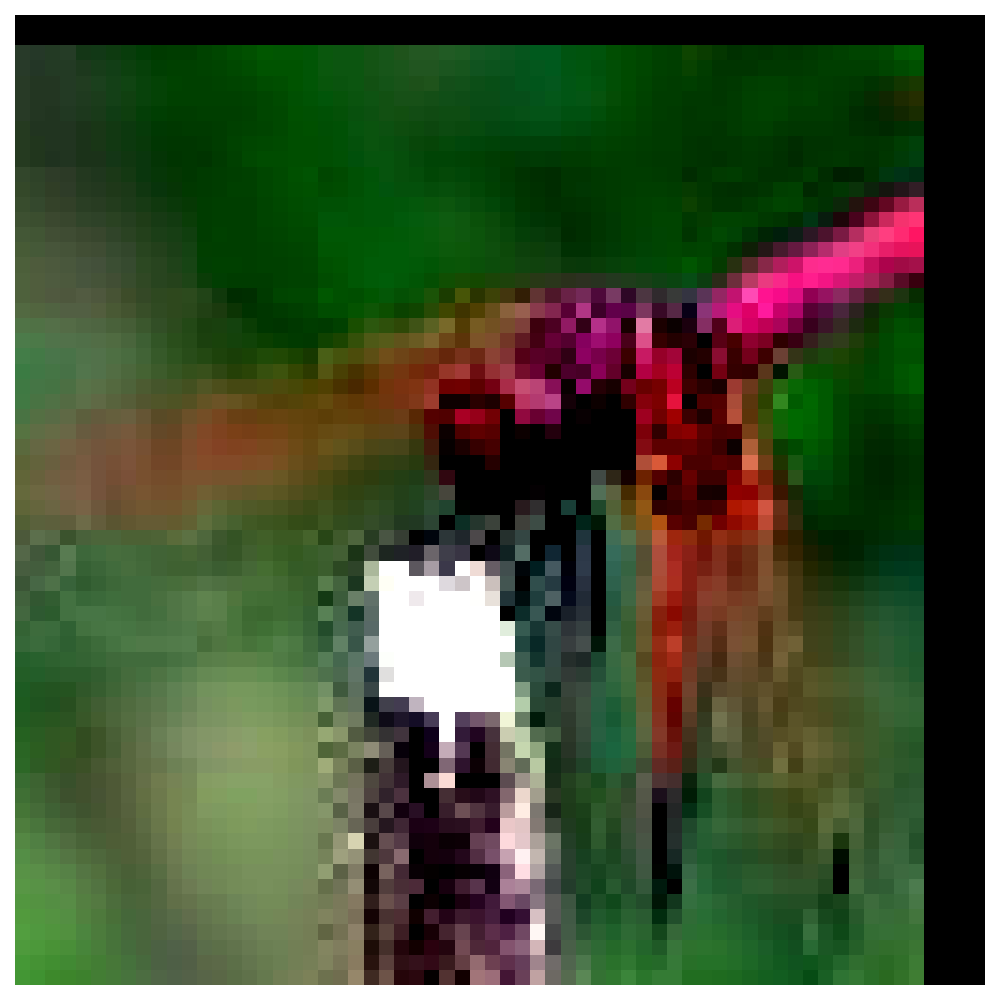}
             \vspace{-1em}
         \end{subfigure}
         
         \begin{subfigure}[b]{\textwidth}
             \centering
             \includegraphics[width=\textwidth]{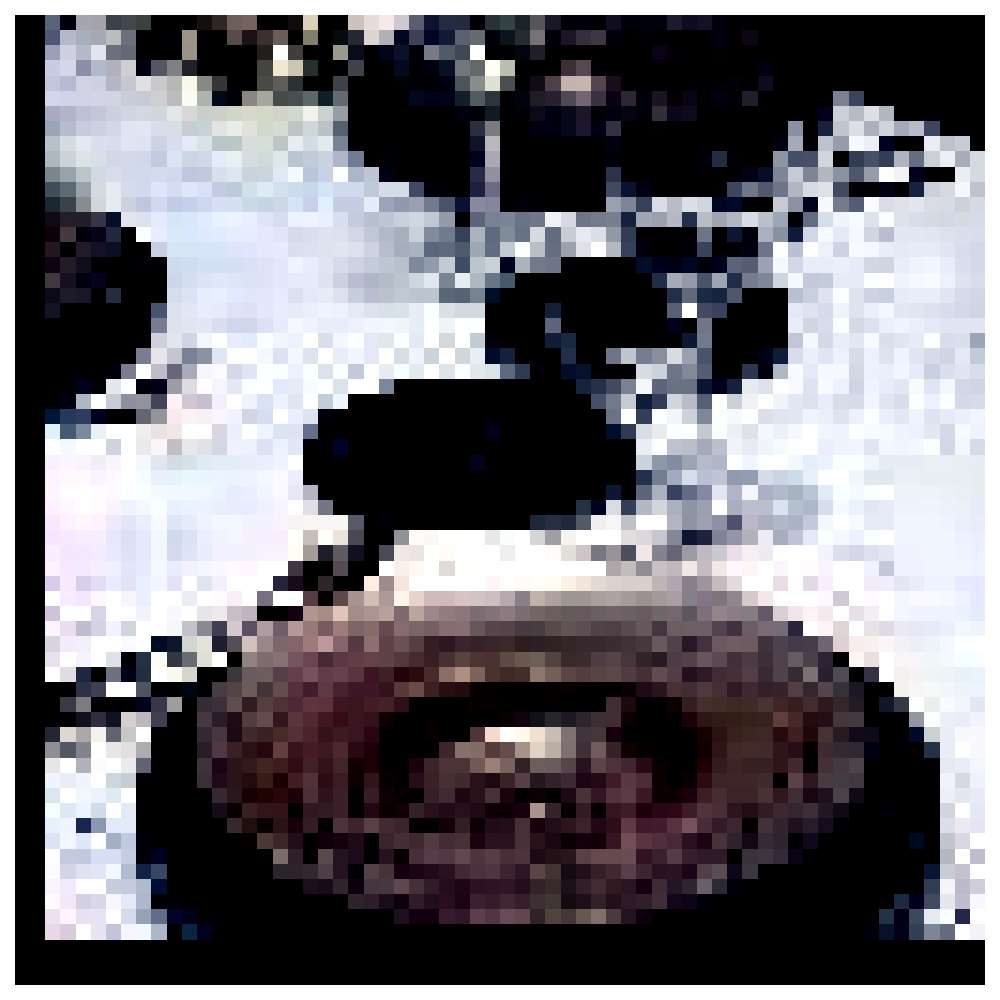}
             \vspace{-1em}
         \end{subfigure}
         
         \begin{subfigure}[b]{\textwidth}
             \centering
             \includegraphics[width=\textwidth]{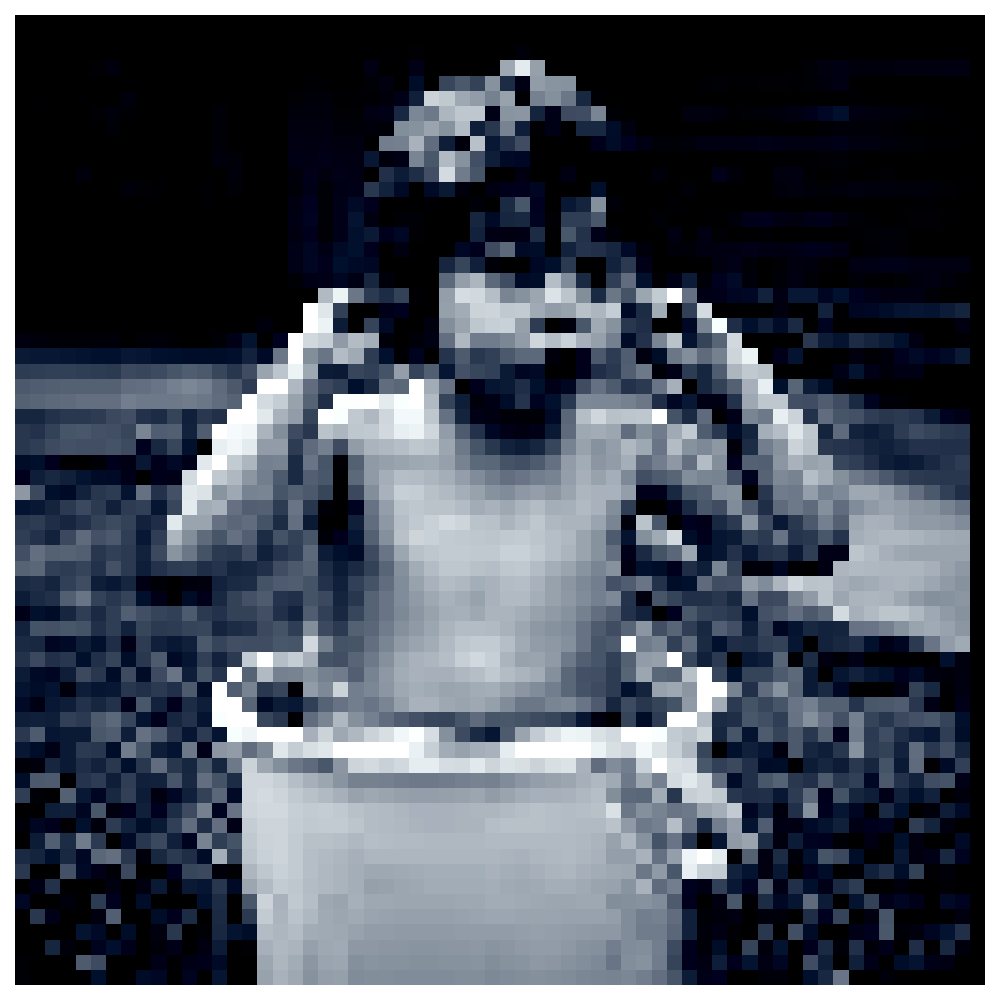}
             \vspace{-1em}
         \end{subfigure}
         \caption{Original image}
    \end{subfigure}
    \begin{subfigure}[b]{0.23\textwidth}
         \centering
         \vspace{-1em}
         \begin{subfigure}[b]{\textwidth}
             \centering
             \hspace{-0.45em}
             \includegraphics[width=\textwidth]{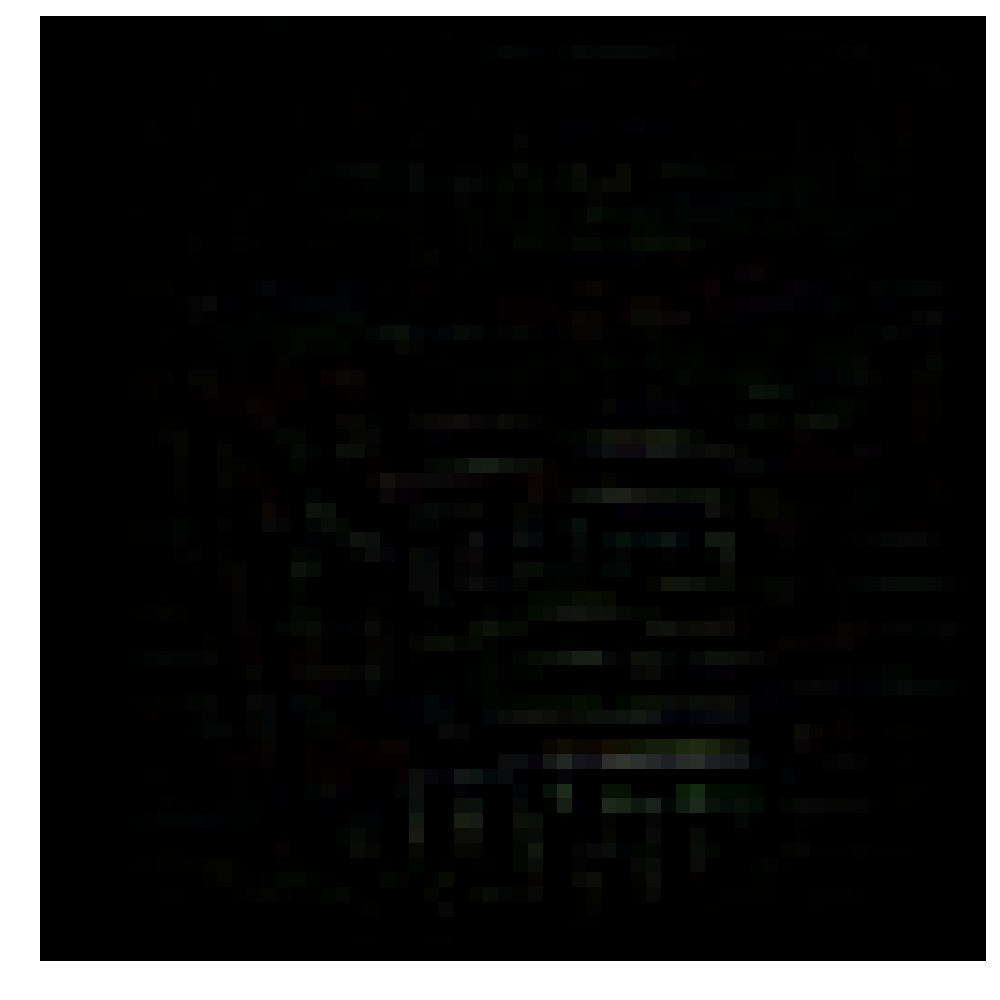}
             \vspace{-1.2em}
         \end{subfigure}

         \begin{subfigure}[b]{\textwidth}
             \centering
             \includegraphics[width=\textwidth]{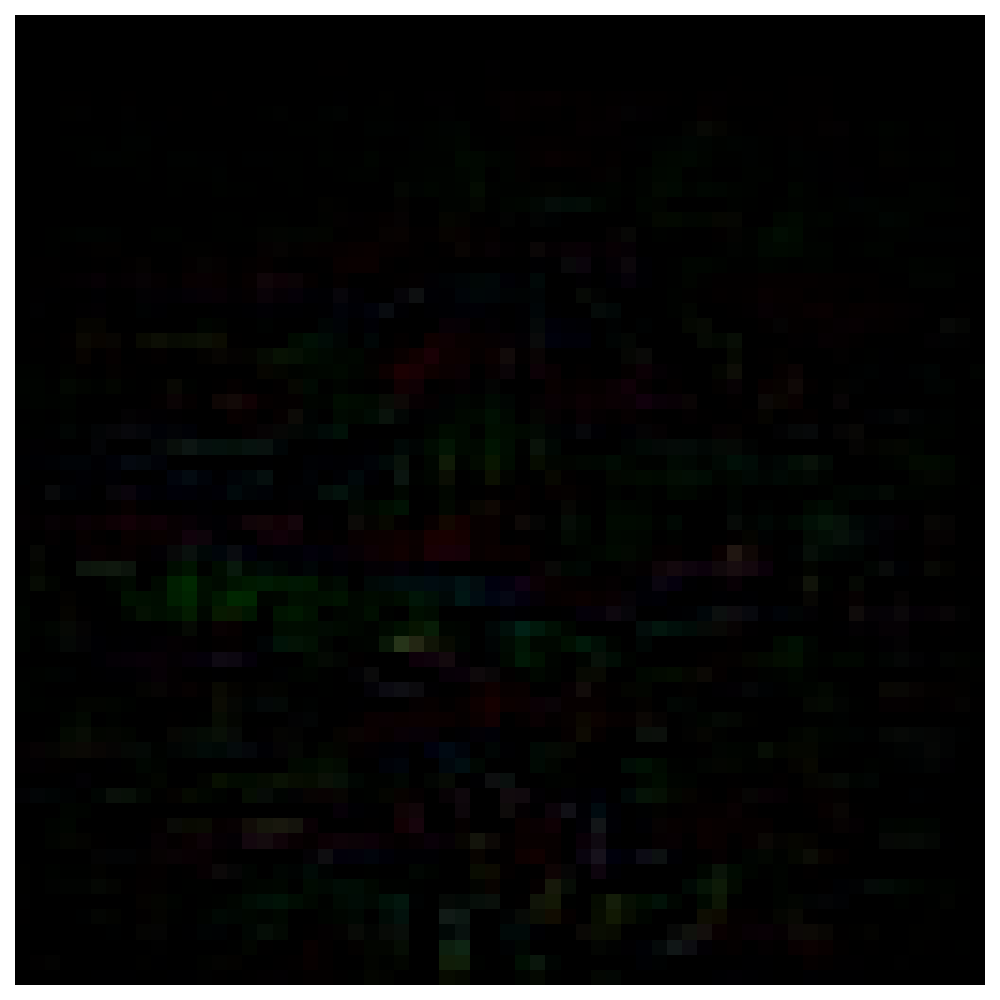}
             \vspace{-1em}
         \end{subfigure}

         \begin{subfigure}[b]{\textwidth}
             \centering
             \includegraphics[width=\textwidth]{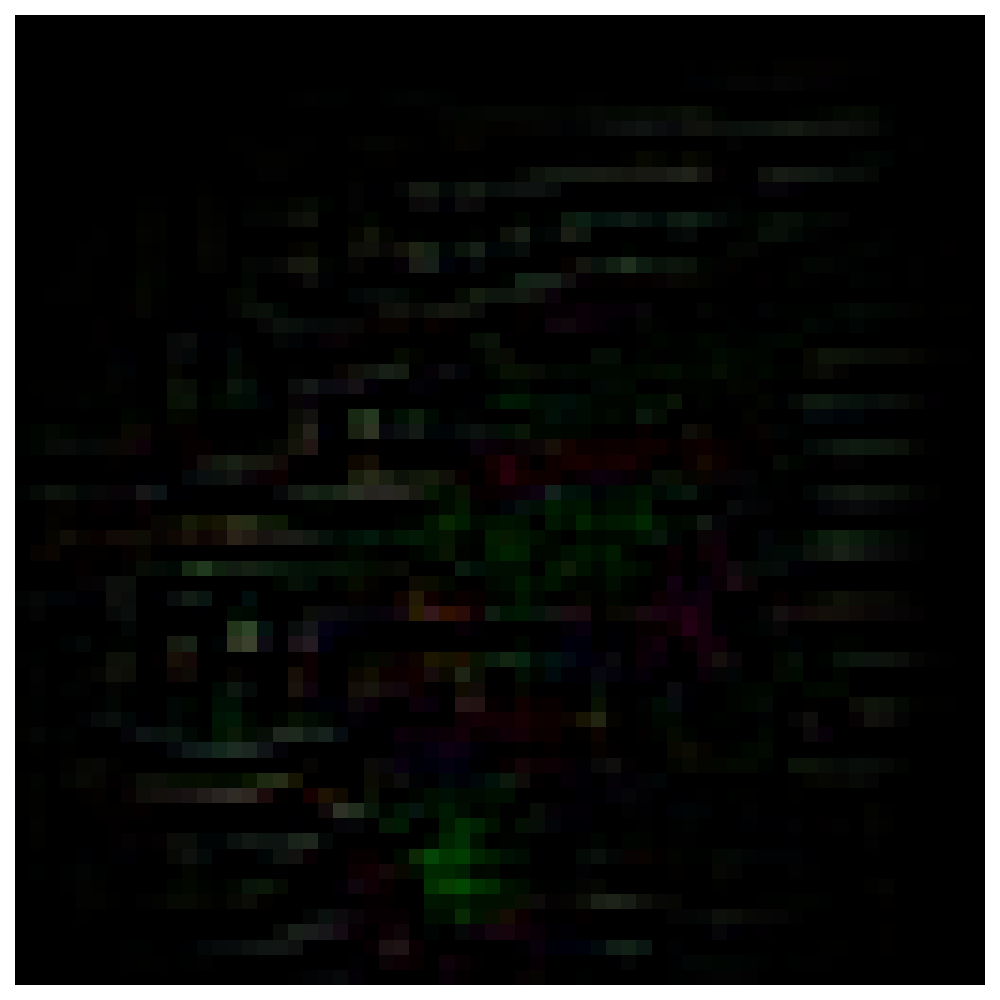}
             \vspace{-1em}
         \end{subfigure}

         \begin{subfigure}[b]{\textwidth}
             \centering
             \includegraphics[width=\textwidth]{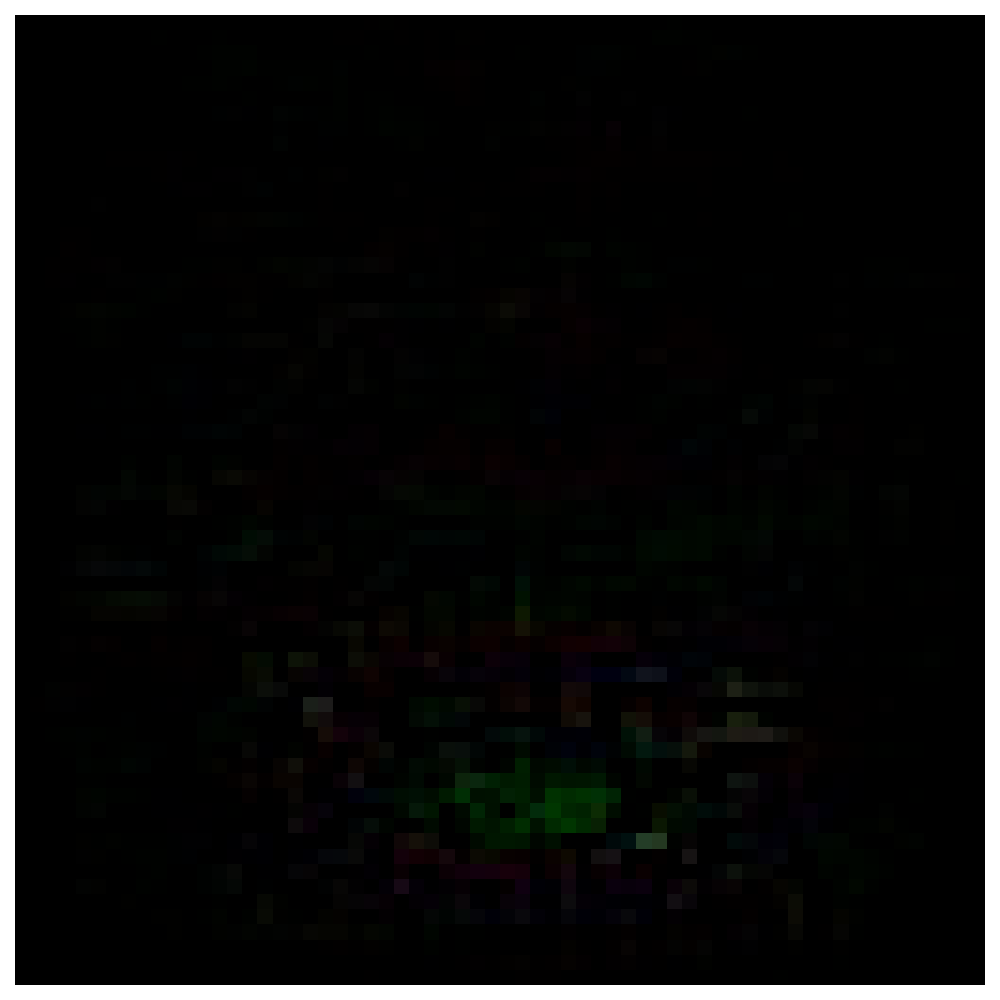}
             \vspace{-1em}
         \end{subfigure}

         \begin{subfigure}[b]{\textwidth}
             \centering
             \includegraphics[width=\textwidth]{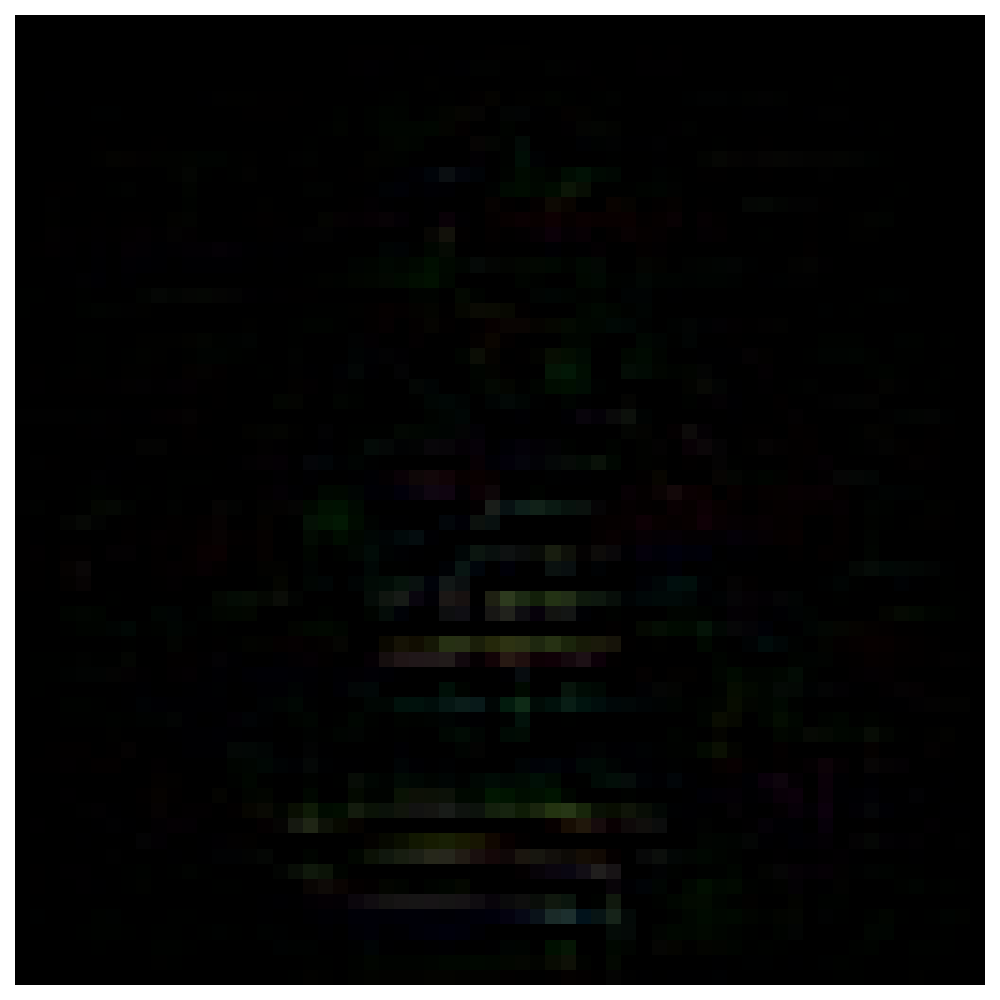}
             \vspace{-1em}
         \end{subfigure}

         \caption{Adversarial perturbation}
    \end{subfigure}
    \begin{subfigure}[b]{0.23\textwidth}
         \centering
         \vspace{-1em}
         \begin{subfigure}[b]{\textwidth}
             \centering
             \hspace{-0.45em}
             \includegraphics[width=\textwidth]{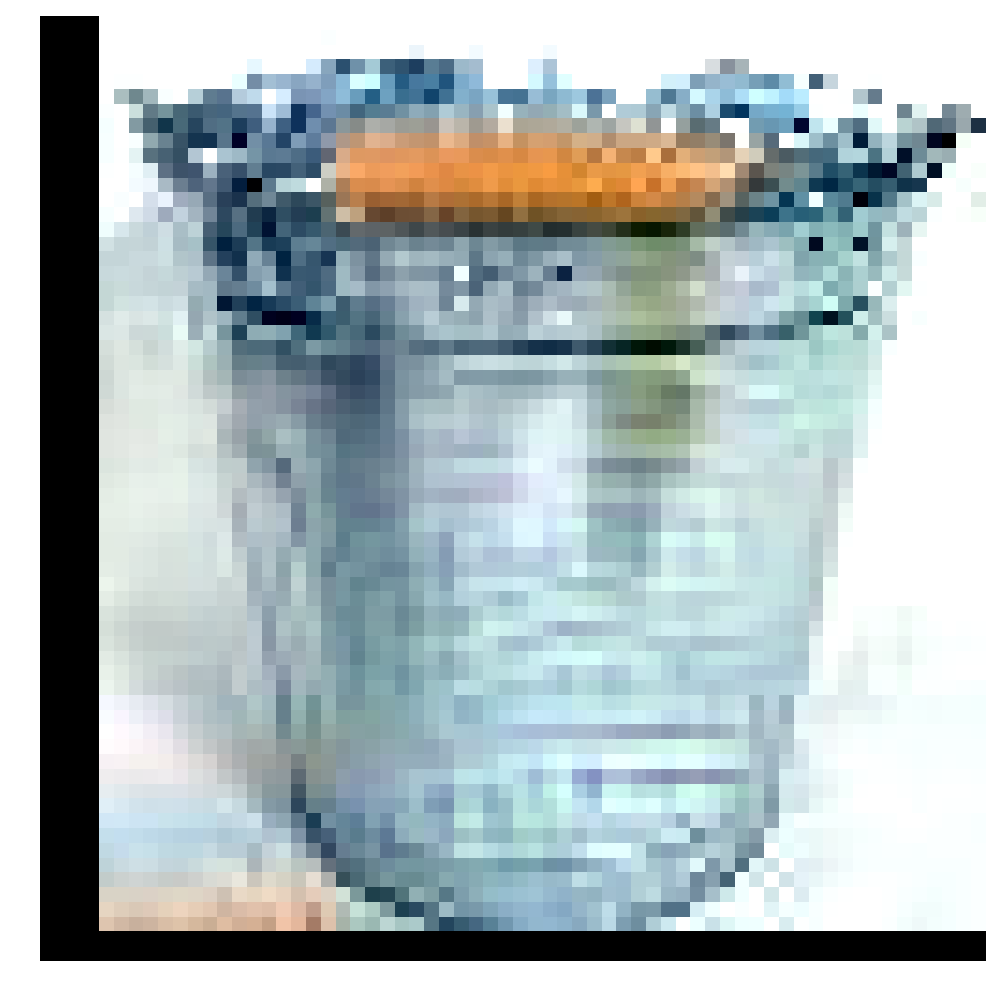}
             \vspace{-1.2em}
         \end{subfigure}

         \begin{subfigure}[b]{\textwidth}
             \centering
             \includegraphics[width=\textwidth]{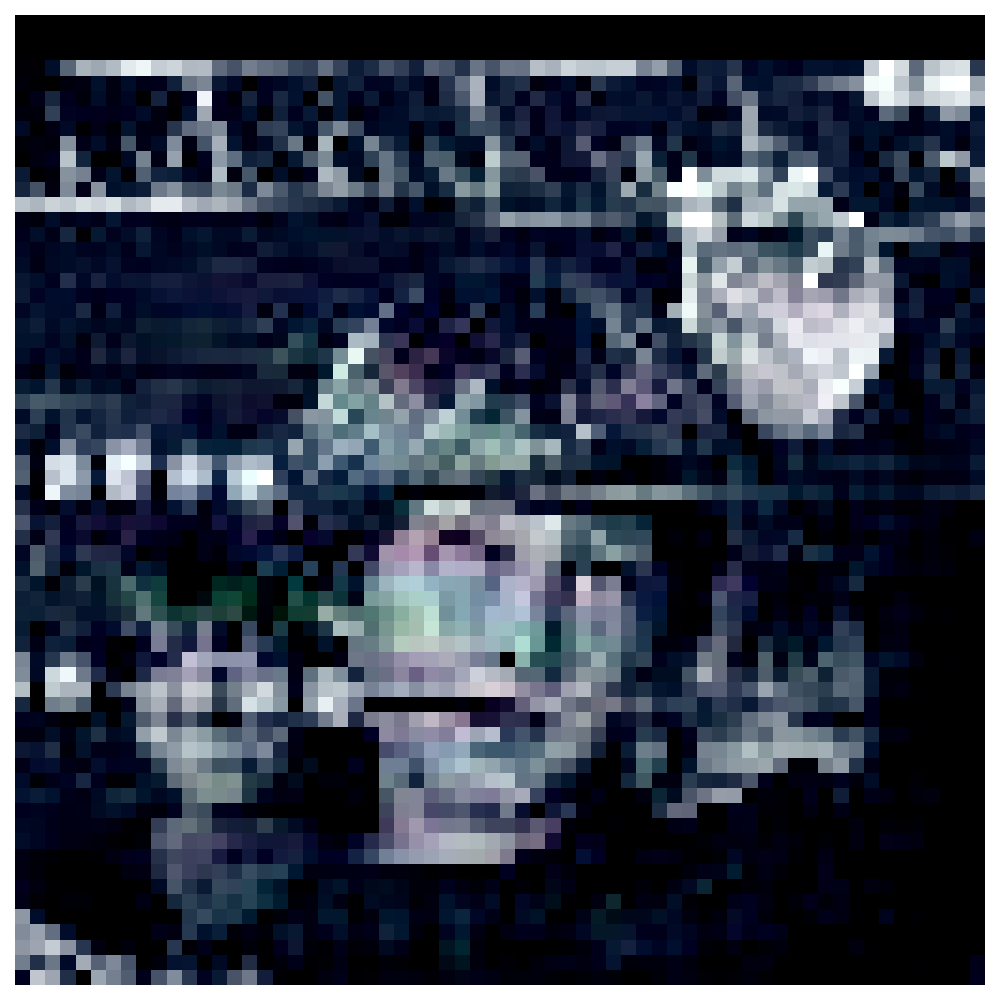}
             \vspace{-1em}
         \end{subfigure}

         \begin{subfigure}[b]{\textwidth}
             \centering
             \includegraphics[width=\textwidth]{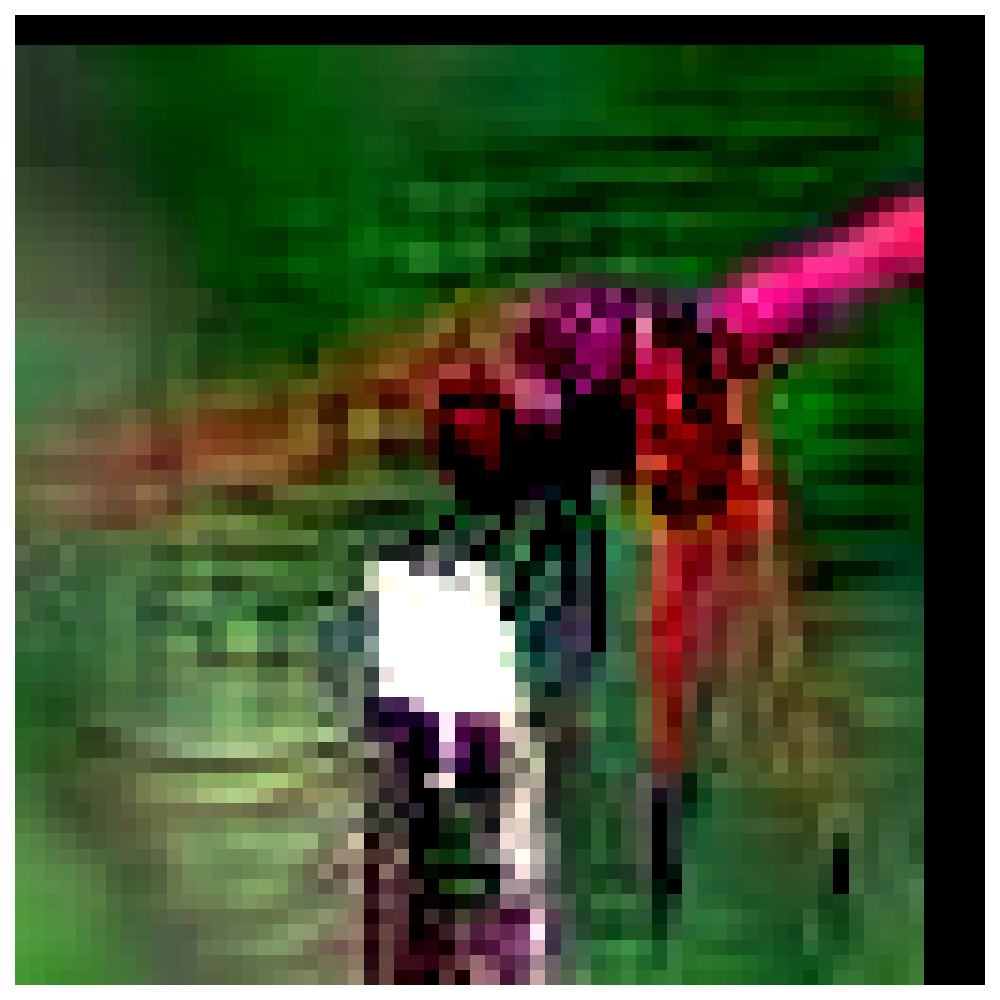}
             \vspace{-1em}
         \end{subfigure}

         \begin{subfigure}[b]{\textwidth}
             \centering
             \includegraphics[width=\textwidth]{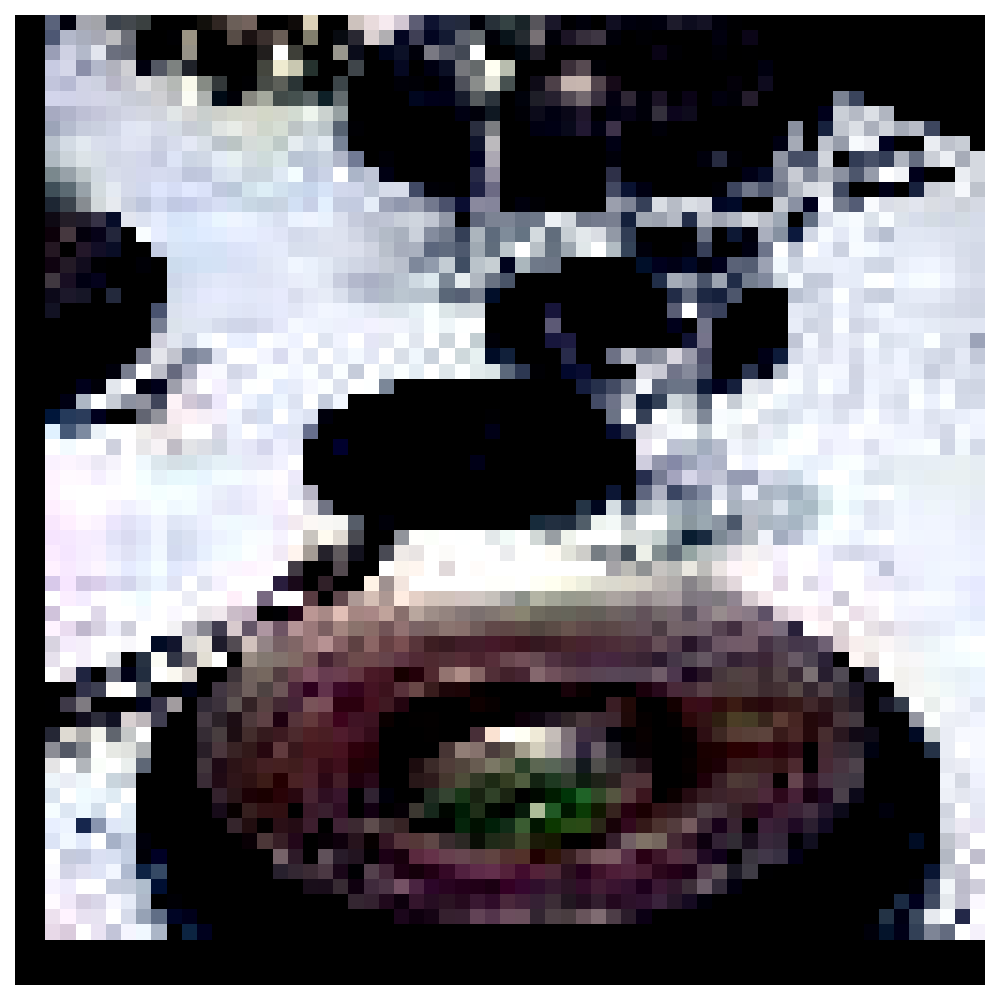}
             \vspace{-1em}
         \end{subfigure}

         \begin{subfigure}[b]{\textwidth}
             \centering
             \includegraphics[width=\textwidth]{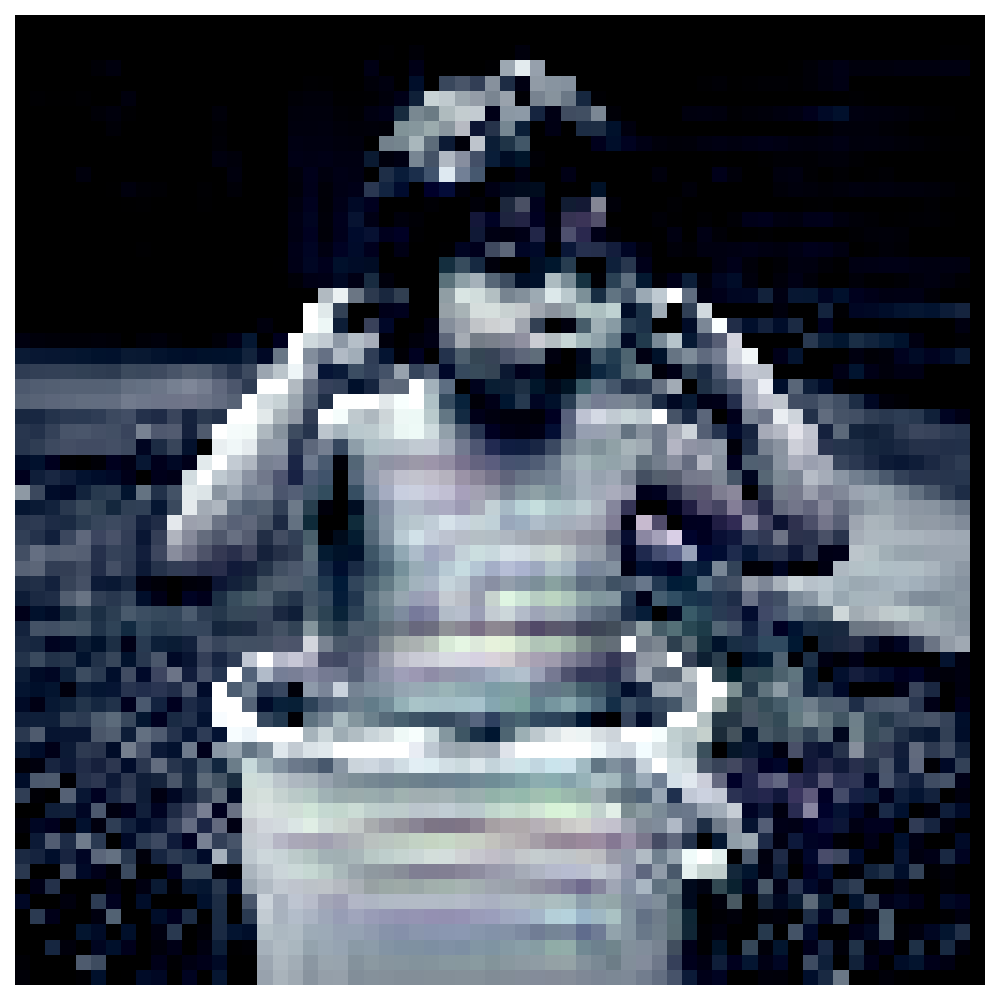}
             \vspace{-1em}
     \end{subfigure}
     \caption{Adversarial image}
    \end{subfigure}
    \caption{Visualization of image, perturbation and the corresponding adversarial image for some samples from TinyImageNet.}
    \label{fig:tiny_combo}
\end{figure*}

We visualize the original and adversarially perturbed images and the corresponding perturbations for CIFAR-100 and TinyImageNet in~\cref{fig:cifar_combo} and~\cref{fig:tiny_combo}. We observe that the perturbations are perceptible in most of the adversarial images generated from low-resolution images of CIFAR-100 and TinyImageNet while for ImageNet-Subset, CUB-200 and Stanford Cars having high-resolution images of 224x224, the perturbations are not perceptible.

{
    \small
    \bibliographystyle{ieeenat_fullname}
    \bibliography{main}
}

\end{document}